\documentclass[10pt,twocolumn,letterpaper]{article}

\usepackage{wacv}              

\usepackage{algorithm}
\usepackage{algpseudocode}
\usepackage{multirow}
\definecolor{wacvblue}{rgb}{0.21,0.49,0.74}
\usepackage[pagebackref,breaklinks,colorlinks,allcolors=wacvblue]{hyperref}

\newtheorem{theorem}{\bf Theorem}

\def\wacvPaperID{2959} 
\def\confName{WACV}
\def\confYear{2027}

\title{LiTe-GS: Oracle-Efficient Next Best View Selection for 3D Gaussian Splatting}

\author{Vivek Pandey\\
\and
Amirhossein Mollaei Khass\\
\and
Nader Motee}

\begin{document}
\maketitle

\begin{abstract}
Selecting informative camera views is critical for efficient training and
adaptive refinement in 3D Gaussian Splatting, where each observation
significantly influences model parameters. However, information-driven
view-selection strategies can require repeated evaluations of expensive
information-gain oracles as the number of candidate views increases.
We propose LiTe-GS, an oracle-efficient method for next best view
selection in 3D Gaussian Splatting. LiTe-GS reduces the number of
information-oracle evaluations by performing randomized subset evaluation
of candidate views rather than exhaustively scoring the full candidate
pool. The resulting approach achieves expected
$O(M\log(1/\epsilon))$ oracle complexity with respect to the number of
candidate views $M$, independent of the selection cardinality $K$, while
providing an explicit trade-off between oracle efficiency and
approximation quality through $\epsilon$.
We provide theoretical guarantees on oracle complexity and approximation
performance under the proposed selection scheme. Experiments on Blender
and Mip-NeRF 360 demonstrate that LiTe-GS maintains reconstruction
quality comparable to Fisher-information-based baselines while
substantially reducing the number of Fisher-oracle evaluations across
different acquisition settings.
\end{abstract}    
\section{Introduction}
\label{sec:intro}

\begin{figure}[t]
\centering

\begin{subfigure}{0.11\textwidth}
    \centering
    \includegraphics[width=\linewidth]{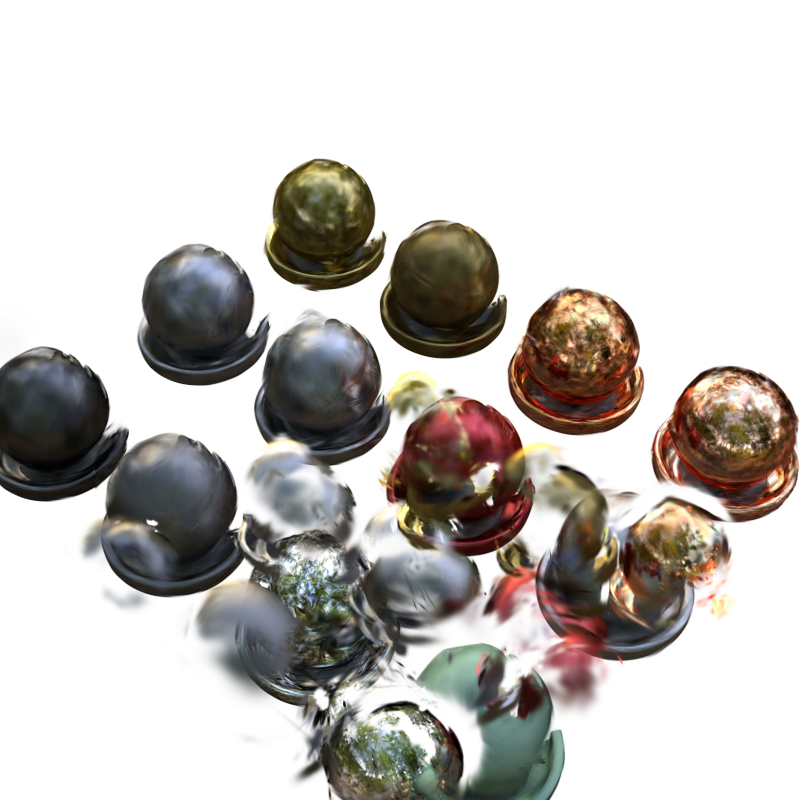}
\end{subfigure}
\hfill
\begin{subfigure}{0.11\textwidth}
    \centering
    \includegraphics[width=\linewidth]{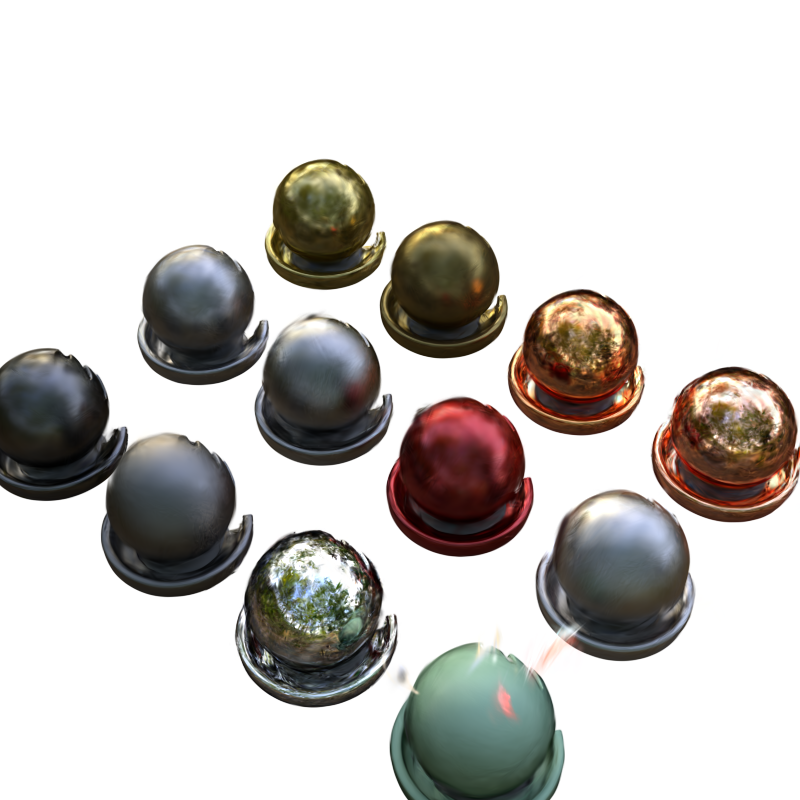}
\end{subfigure}
\hfill
\begin{subfigure}{0.11\textwidth}
    \centering
    \includegraphics[width=\linewidth]{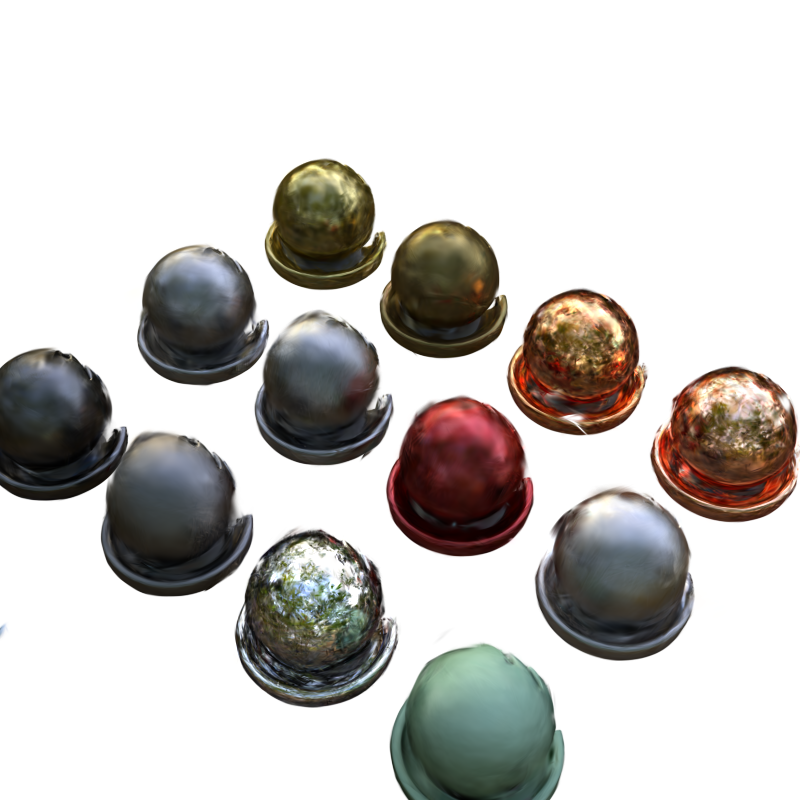}
\end{subfigure}
\hfill
\begin{subfigure}{0.11\textwidth}
    \centering
    \includegraphics[width=\linewidth]{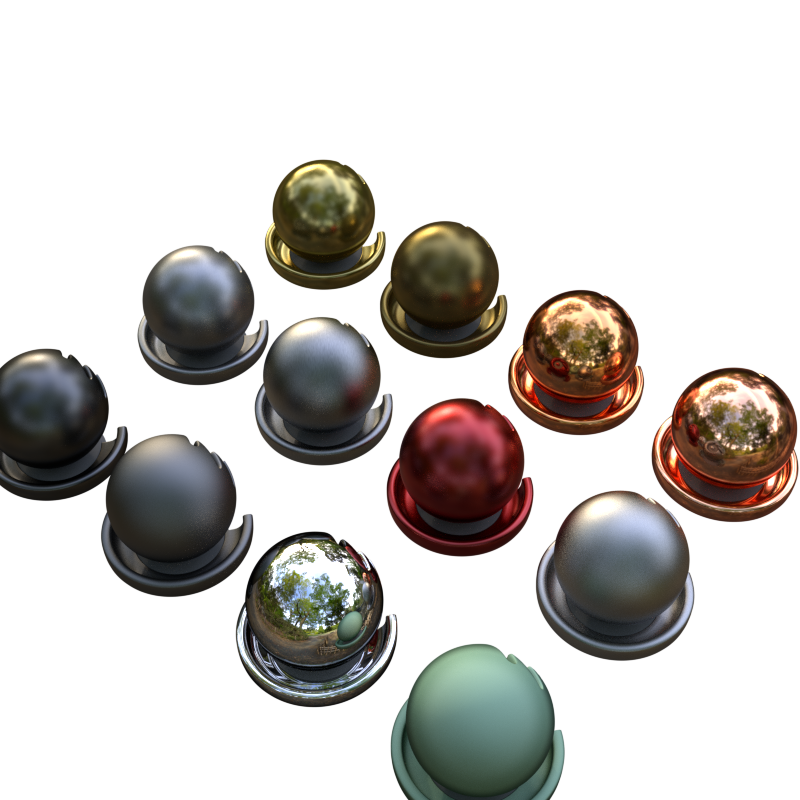}
\end{subfigure}

\vspace{0.11cm}

\begin{subfigure}{0.11\textwidth}
    \centering
    \includegraphics[width=\linewidth]{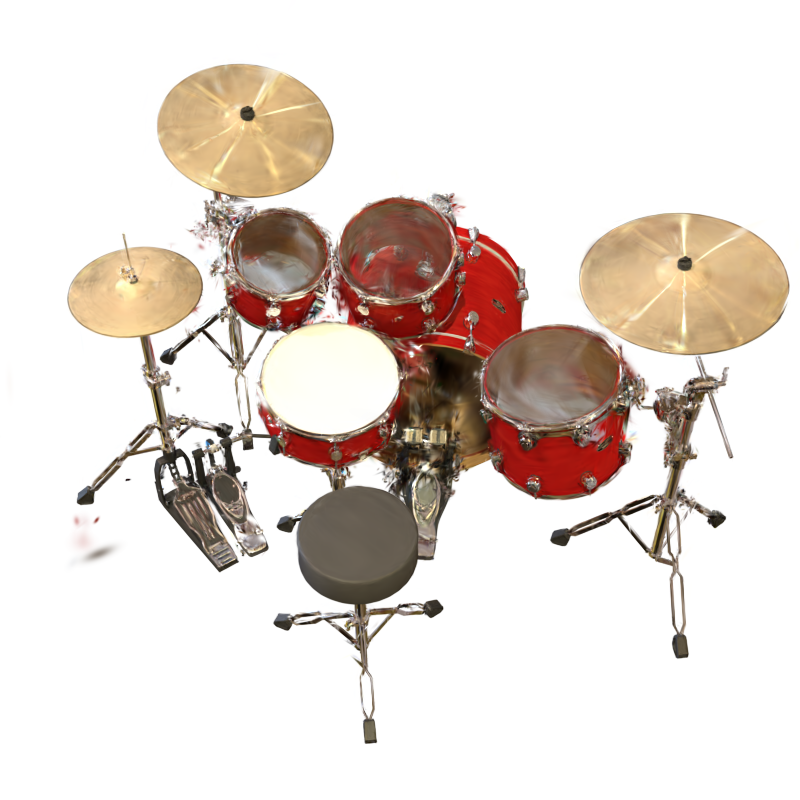}
\end{subfigure}
\hfill
\begin{subfigure}{0.11\textwidth}
    \centering
    \includegraphics[width=\linewidth]{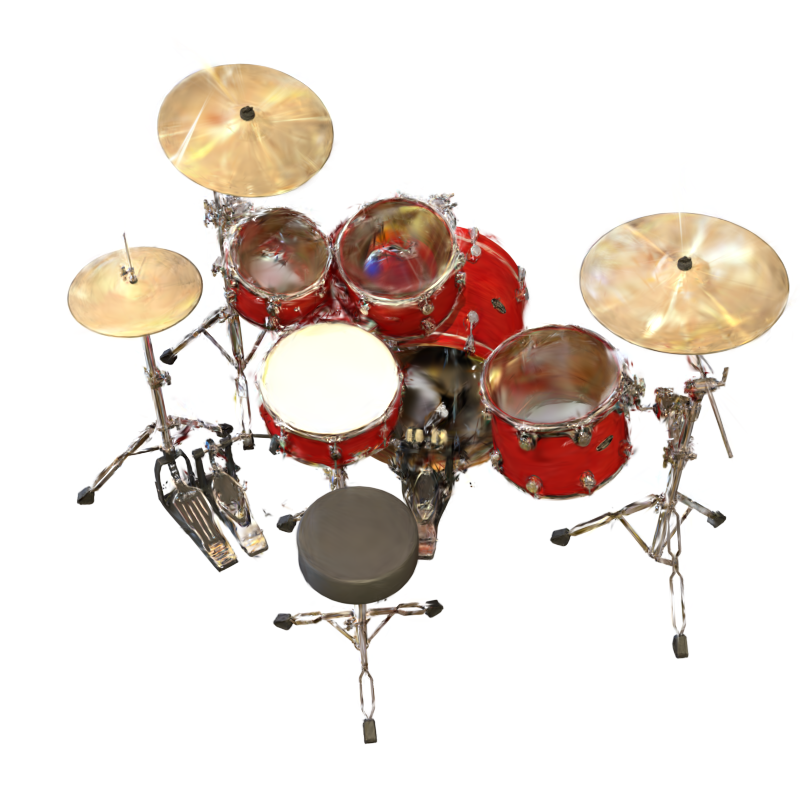}
\end{subfigure}
\hfill
\begin{subfigure}{0.11\textwidth}
    \centering
    \includegraphics[width=\linewidth]{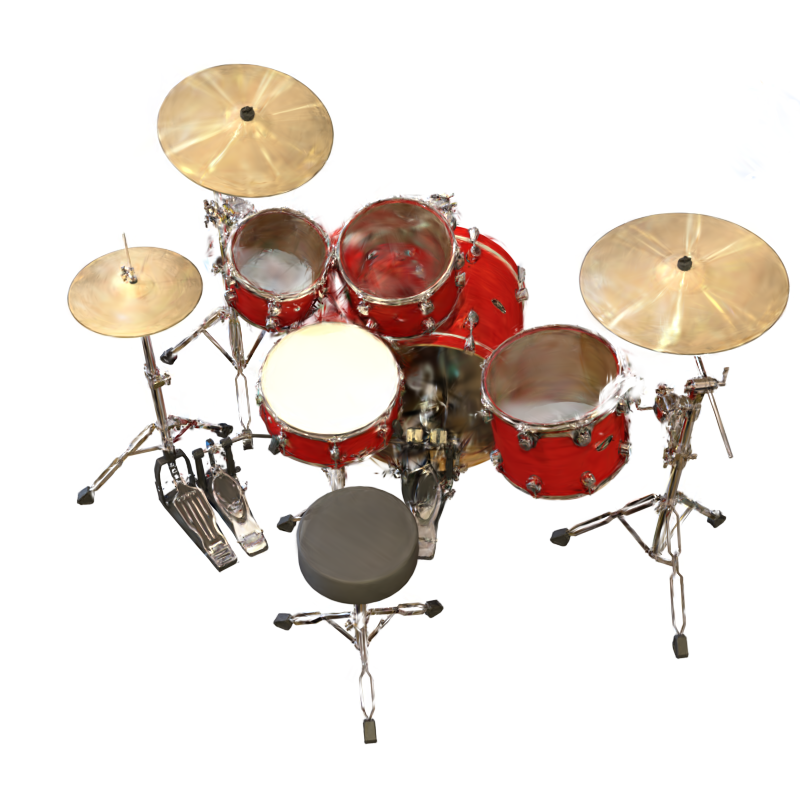}
\end{subfigure}
\hfill
\begin{subfigure}{0.11\textwidth}
    \centering
    \includegraphics[width=\linewidth]{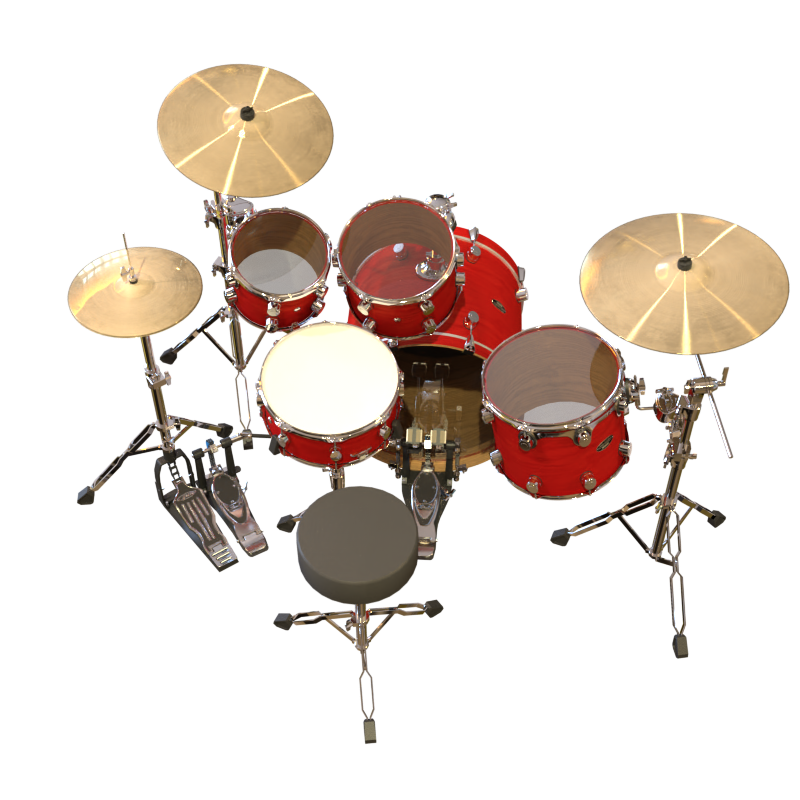}
\end{subfigure}

\vspace{-0.11cm}

\begin{subfigure}{0.11\textwidth}
    \centering
    \includegraphics[width=\linewidth]{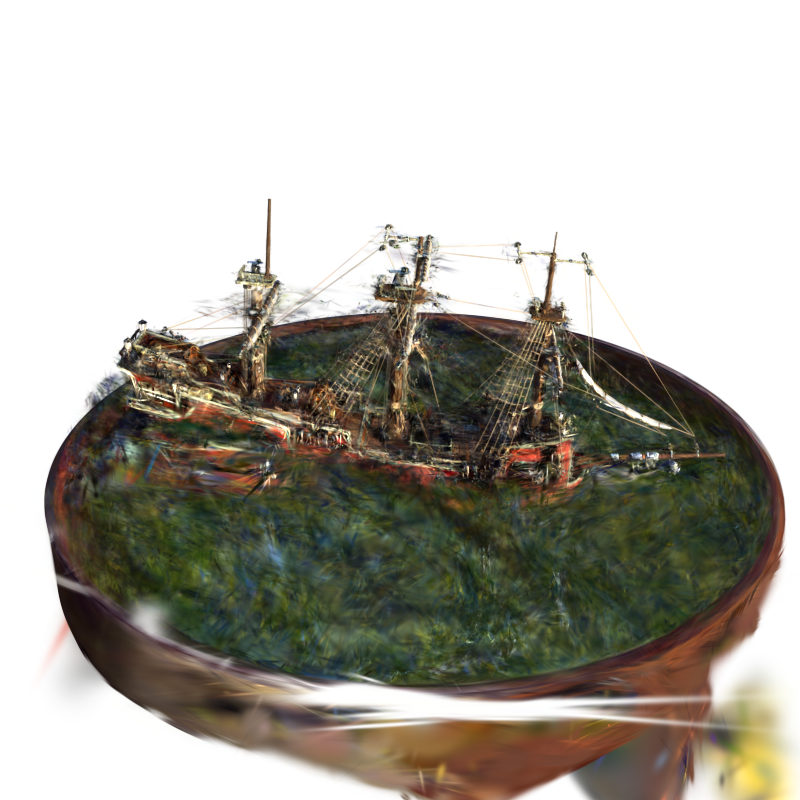}
    \caption{Random}
\end{subfigure}
\hfill
\begin{subfigure}{0.11\textwidth}
    \centering
    \includegraphics[width=\linewidth]{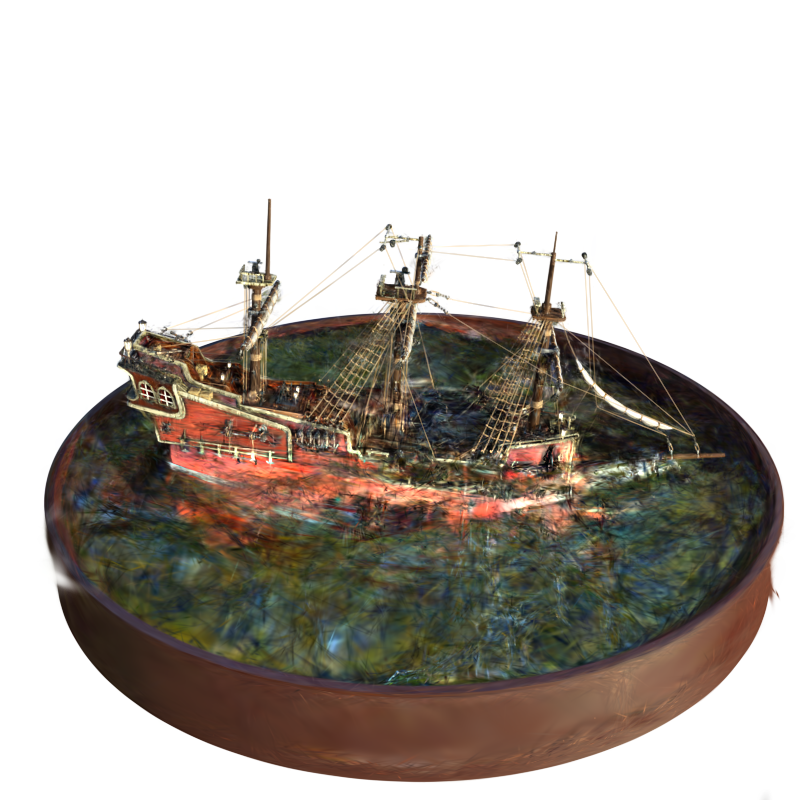}
    \caption{POP-GS}
\end{subfigure}
\hfill
\begin{subfigure}{0.11\textwidth}
    \centering
    \includegraphics[width=\linewidth]{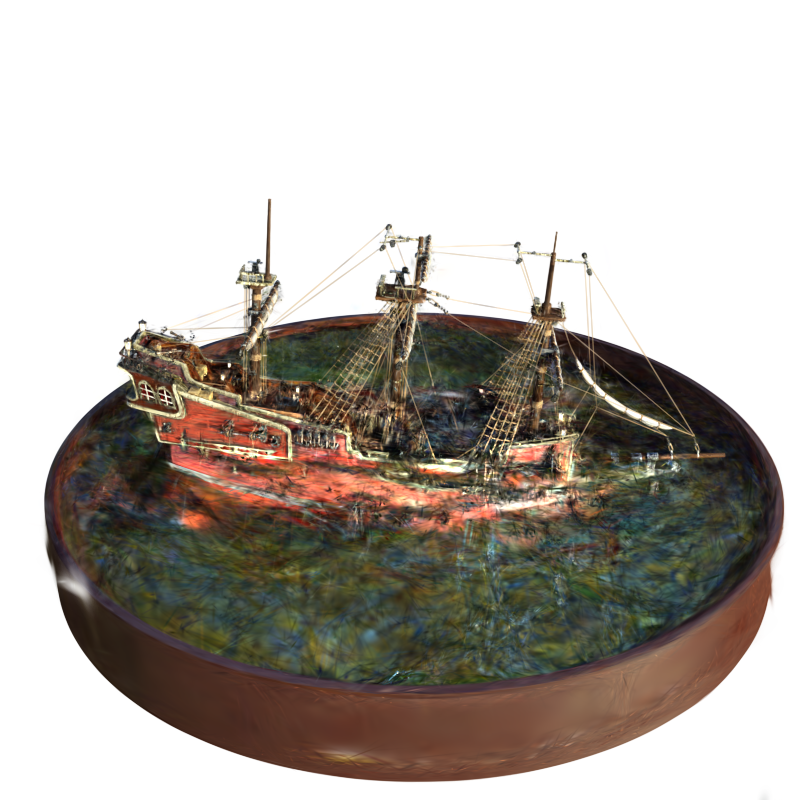}
    \caption{Ours}
\end{subfigure}
\hfill
\begin{subfigure}{0.11\textwidth}
    \centering
    \includegraphics[width=\linewidth]{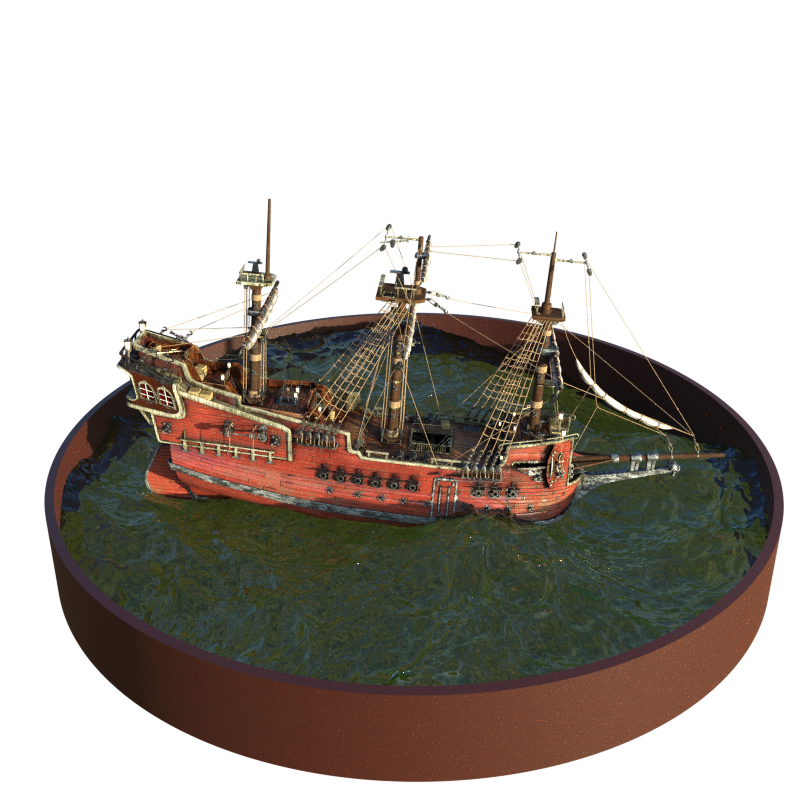}
    \caption{Ground Truth}
\end{subfigure}
\caption{Setup 1: Qualitative comparison of rendered images on the Blender dataset. Results obtained using Random sampling, POP-GS~\cite{wilson2025pop}, and LiTe-GS are shown alongside the corresponding ground-truth images.}
\label{fig:blender_exp1}
\vspace{-0.4cm}

\end{figure}



Volumetric rendering has become a fundamental paradigm for photorealistic novel-view synthesis in computer vision and robotics, enabling scene reconstruction by modeling light transport along camera rays. Recent advances have enabled high-quality reconstruction from multi-view imagery using both implicit and explicit scene representations. Neural Radiance Fields (NeRF)~\cite{mildenhall2022nerf} demonstrated high-fidelity reconstruction using implicit neural scene parameterizations, while subsequent works such as Instant Neural Graphics Primitives~\cite{muller2022instant}, Plenoxels~\cite{yu2022plenoxels}, and Mip-NeRF 360~\cite{barron2022mip} improved scalability and optimization efficiency through accelerated volumetric representations. Despite these advances, volumetric rendering methods remain computationally demanding for latency-sensitive applications common in robotics and real-time perception tasks such as autonomous navigation and simultaneous localization and mapping (SLAM).

\begin{figure*}
  \centering
        \centering
    \includegraphics[width=\linewidth]{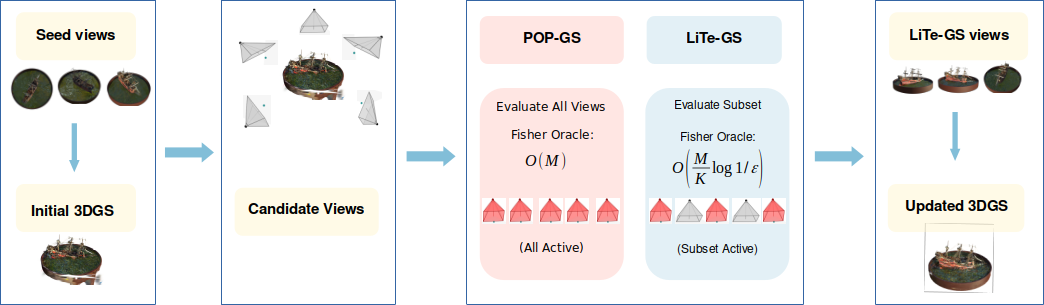}
    \caption{Overview of the LiTe-GS pipeline. Given a small set of seed views, we first build an initial 3D Gaussian Splatting (3DGS) reconstruction and define a pool of $M$ candidate viewpoints. We then perform iterative view selection ($K$ views) to refine the reconstruction. We compare POP-GS \cite{wilson2025pop}, which evaluates the full Fisher oracle over all candidates, with LiTe-GS, which uses an $\varepsilon$-constrained randomized strategy that evaluates only a subset of views for efficient selection.
The pipeline proceeds in four stages: seed views and initial reconstruction, candidate view generation, view selection, and final refinement using the selected views.}
    \label{fig:pipeline}
\end{figure*}

The seminal work on 3D Gaussian Splatting (3DGS)~\cite{kerbl3Dgaussians} introduced a fundamentally explicit and anisotropic representation using millions of 3D Gaussians-each with optimized position, orientation, scaling, opacity, and spherical harmonics-paired with a fast, differentiable, visibility-aware rasterization pipeline. By avoiding volumetric sampling in empty space and leveraging efficient GPU splatting, 3DGS achieves reconstruction quality comparable to or exceeding NeRF and Plenoxels while drastically reducing training time and enabling real-time rendering. These properties make 3DGS particularly attractive for robotics applications \cite{zhu20243dgaussiansplattingrobotics}, where efficient, high-fidelity 3D scene modeling from sequential or multi-view observations supports tasks such as real-time mapping and SLAM~\cite{keetha2024splatam,yan2024gs, matsuki2024gaussian, carlone2025slam}, obstacle avoidance and risk aware navigation~\cite{liu2024beyond,chen2025splat,khass2025active}, and robotic manipulation~\cite{wu2025rl, swann2024touch}.
By providing rich geometric and appearance information at low latency, 3DGS can enhance spatial understanding in autonomy pipelines.

Despite its efficiency, accurate scene reconstruction with 3D Gaussian Splatting (3DGS) \cite{kerbl3Dgaussians} still requires a sufficiently informative set of training images selected from a potentially large pool of candidate viewpoints. In practical view acquisition settings, processing all available views is computationally expensive and often introduces redundant observations with limited additional information gain. Consequently, view selection becomes critical for identifying frames that most effectively reduce uncertainty in the underlying 3DGS model parameters while maintaining computational efficiency.

Recent works have explored information-driven view selection and next-best-view strategies for training Gaussian splatting models. FisherRF~\cite{Jiang2023FisherRF} formulates view selection through information gain maximization, while POP-GS~\cite{wilson2025pop} extends this perspective using principles from optimal experimental design. 
Despite these advances, information-driven view selection still entails a
trade-off between the number of candidate-oracle evaluations and the
quality of the resulting selection. Exhaustive greedy evaluation queries
the full candidate set at each acquisition step, whereas reducing the
number of evaluations generally requires restricting the candidates
considered. This motivates an oracle-efficient selection strategy that
controls this trade-off explicitly through randomized candidate sampling.


Building upon the foundations of FisherRF~\cite{Jiang2023FisherRF} and POP-GS~\cite{wilson2025pop}, we introduce \textit{LiTe-GS}, an oracle-efficient next best view selection framework that balances efficiency and information gain while providing provable approximation guarantees.

\vspace{0.11cm}





\noindent Our main contributions are:
\begin{itemize}
\item We formulate active next best view (NBV) selection in 3DGS as a binary optimization problem, explicitly linking view selection to uncertainty reduction in model parameters.

\item We propose \textit{LiTe-GS}, an oracle-efficient view selection algorithm that reduces the number of Fisher oracle evaluations through randomized subset selection.


\item We provide theoretical analysis characterizing the oracle
complexity and approximation guarantees of the proposed algorithm, as
well as the efficiency-approximation quality trade-off
controlled by $\epsilon$.
\end{itemize}

\section{Related Works}
\label{sec:related_works}
This section reviews recent advances in 3D Gaussian Splatting (3DGS), followed by active view selection and next-best-view planning methods.

\subsection{3D Gaussian Splatting}

Recent advances in novel view synthesis have led to a variety of neural scene representations, including implicit radiance fields \cite{mildenhall2022nerf,barron2022mip}, voxel-based methods \cite{yu2022plenoxels}, point-based representations \cite{xu2022pointnerfCVPR}, and hybrid approaches \cite{sun2022directvoxelCVPR,muller2022instant}. While these methods achieve strong rendering quality, many incur high training and rendering costs.

Recently, 3D Gaussian Splatting (3DGS) \cite{kerbl3Dgaussians} has emerged as an efficient explicit representation for novel view synthesis \cite{meuleman2025onfly,NEURIPS2024_bb11f79a,paliwal2024coherentgs}. Unlike implicit radiance field methods, 3DGS represents a scene as a set of anisotropic 3D Gaussians parameterized by position, covariance, opacity, and appearance attributes. The representation is optimized via differentiable rasterization
enabling high-quality reconstruction with real-time rendering performance.


The efficiency and explicit geometric structure of 3DGS have also motivated research in robotics  \cite{fei2024survey}, particularly in SLAM \cite{zhu2023nice_slam,matsuki2024gaussian,carlone2025slam}. Several works integrate Gaussian representations into SLAM pipelines: Photo-SLAM \cite{Huang_2024_phoslam_CVPR} combines Gaussians with ORB-SLAM, while GS-ICP SLAM \cite{Ha_2024_ECCV} and RTG-SLAM \cite{peng2024rtgslam} leverage ICP-based registration for robust tracking and mapping. In addition, recent approaches incorporate LiDAR measurements into Gaussian-based mapping frameworks \cite{Sun_2024_IROS,Lang_2025_ICRA,Hong_2024_LIV_GaussMap}, demonstrating the adaptability of Gaussian representations across sensing modalities.

While most existing works assume a fixed set of input views for training and focus on downstream tasks, view selection for efficient Gaussian reconstruction remains less explored. This work considers this direction by exploring an uncertainty-aware view selection strategy for 3D Gaussian Splatting.


\subsection{Active View Selection and Next-Best-View Planning}
Driven by the need to reduce training costs and enable active view acquisition for 3D Gaussian Splatting (3DGS), recent works have explored selecting informative viewpoints based on cues extracted from trained models, existing observations, and candidate views.

A central theme across these methods is either black box evaluation \cite{polyzos2025activeinitsplatactiveimageselection} or the estimation and exploitation of uncertainty \cite{wilson2025modeling,xue2026uncertainty}.
FisherRF~\cite{Jiang2023FisherRF} formulates next-best-view selection as maximizing expected information gain derived from the Fisher Information Matrix (FIM) over candidate views. However, explicitly storing the full Fisher information matrix is impractical due to memory constraints. To address this, FisherRF adopts a diagonal approximation of the FIM, improving efficiency while trading off approximation fidelity.
POP-GS~\cite{wilson2025pop} further develops this formulation using a block-diagonal approximation,  
improving estimation accuracy while maintaining tractability.
More recently, \cite{li2026next} extends next-best-view selection to dynamic 3D Gaussian Splatting settings.

Despite these advances, candidate-view selection still requires repeated
Fisher-information evaluations across the candidate set. This motivates
an oracle-efficient selection strategy that reduces the number of
evaluations while retaining the information-driven objective and exposing
a direct trade-off between oracle expenditure and approximation quality.
To this end, we propose a lightweight active next-best-view selection
framework, LiTe-GS.

\section{View Selection Formulation}\label{sec:view_selection_formulation}
We consider the problem of informative view selection for efficient training in 3D Gaussian Splatting (3DGS). Given a large pool of candidate viewpoints, the problem is to select an informative subset of views in an efficient manner. We formalize this as an uncertainty reduction problem in the estimation of 3DGS parameters.

We first briefly review the parameterized representation used in 3D Gaussian Splatting.

\subsection{3D Gaussian Splatting}
3D Gaussian Splatting represents a scene using a collection of anisotropic
Gaussian primitives. Let
\(
\mathcal{G} = \{ g_i \}_{i=1}^{N}
\)
denote a set of $N$ Gaussians, where each Gaussian is parameterized by
\begin{equation}
\theta_i =
\{\mu_i, q_i, s_i, \alpha_i, c_i\},
\end{equation}
where
$\mu_i \in \mathbb{R}^3$ denotes the 3D mean position,
$q_i \in {S}^3$ represents rotation using a unit quaternion,
$s_i \in \mathbb{R}^3$ denotes anisotropic scaling parameters,
$\alpha_i \in \mathbb{R}$ is the opacity, and
$c_i \in \mathbb{R}^{3SH}$ corresponds to spherical harmonic (SH)
coefficients encoding view-dependent RGB radiance.

Collectively, the parameters of the scene are given by
\begin{equation}
\Theta = \{\theta_i\}_{i=1}^{N}.
\end{equation}
Given a set of training views \(
\mathcal{V},
\)
3DGS estimates $\Theta$ through gradient based optimization as follows: 
\begin{equation}
\hat{\Theta}
=
\arg\min_{\Theta}
\sum_{v \in \mathcal{V}}
\mathcal{L}(v;\Theta),
\end{equation}
where $\mathcal{L}(\cdot,\cdot) = (1 - \lambda) \mathcal{L}_1 + \lambda \mathcal{L}_{SSIM}$ denotes the weighted sum of $\mathcal{L}_1$ loss and SSIM loss between rendered image and ground truth image.

Each 3D Gaussian defines a density over 3D space as
\begin{equation}\label{eq:3dgs_density}
    G_i(\mathbf{x}) =
\exp\!\left(
-\tfrac{1}{2}
(\mathbf{x}-\mu_i)^\top \Pi_i^{-1} (\mathbf{x}-\mu_i)
\right),
\end{equation}
where $\mathbf{x} \in \mathbb{R}^3$ is a spatial point and $\Pi_i \in \mathbb{S}_+^3$ is the covariance matrix controlling the spatial extent and orientation of the Gaussian.

For differentiable optimization, the covariance is parameterized as
\begin{equation}\label{eq:3dgs_covariance}
    \Pi_i = R_i S_i S_i^\top R_i^\top,
\end{equation}
where $R_i \in SO(3)$ is a rotation matrix and $S_i$ is a scaling matrix.

Under a camera transformation, the covariance is mapped to image space via
\begin{equation}\label{eq:3dgs_covaraince_image_space}
    \Pi_i' = J W \Pi_i W^\top J^\top,
\end{equation}
where $W$ is the viewing transformation and $J$ is the Jacobian of the projective transformation.

Finally, the rendered color at a pixel is obtained through alpha blending of ordered Gaussian contributions:
\begin{equation}\label{eq:3dgs_alpha_blending}
    C = \sum_{i \in \mathcal{N}} c_i \alpha_i \prod_{j < i} (1 - \alpha_j),
\end{equation}
where $\mathcal{N}$ denotes the set of Gaussians along the pixel’s viewing ray (i.e., those that project onto the pixel), and $c_i$ and $\alpha_i$ denote the color and opacity induced by the projected Gaussian in image space.

\subsection{Uncertainty Reduction in Active View Selection}

We consider an active next best view selection setting where images are not
initially available and observations are acquired only after selecting
viewpoints from a candidate set
$\mathcal{V}$.
An initial set of seed views
$\mathcal{V}_0 \subset \mathcal{V}$
is first used to obtain a prior estimate of the 3DGS parameters
$\hat{\Theta}$.
Assuming locally Gaussian behavior around the optimum, the
posterior covariance can be approximated \cite{bishop2006prml} as
\begin{equation}\label{eqn:posterior_covariance}
\Sigma({\Theta})
\approx
\left( H({\hat{\Theta}}) + \Sigma_0^{-1} \right)^{-1},
\end{equation}
where $\Sigma_0$ denotes prior uncertainty obtained after initial
training and $H({\hat{\Theta}})$  corresponds to the Fisher Information of the selected views.

The rendered observation from view $v$ is modeled as
\begin{equation}\label{eqn:rendering_model}
\mathbf{y}_v = f_v(\Theta) + \eta_v, \quad \eta_v \sim \mathcal{N}(0, \sigma^2 I),
\end{equation}
where $f_v(\Theta)$ denotes the differentiable rendering function induced by the 3D Gaussian representation described above, and $\eta_v$ models isotropic measurement noise.


Assuming a local linearization around the current estimate $\hat{\Theta}$, we have
\begin{equation}
f_v(\Theta) \approx f_v(\hat{\Theta}) + J_v (\Theta - \hat{\Theta}),
\end{equation}
where $J_v = \frac{\partial f_v}{\partial \Theta} \big|_{\hat{\Theta}}$ is the Jacobian of the rendering model with respect to the Gaussian parameters.

The Fisher Information of a set of views $\bar{\mathcal{V}} \subseteq \mathcal{V}$ is calculated as
\begin{equation}
H_{\bar{\mathcal{V}}}({\hat{\Theta}})
\approx \sigma^{-2}
\sum_{v \in {\bar{\mathcal{V}}}}
J_v^\top J_v,
\end{equation}
where $J_v$ denotes the Jacobian of rendered observations with respect
to the Gaussian parameters for view $v$.


However, the full information matrix scales with the total number of Gaussian parameters, which can reach millions in large scenes,
resulting in prohibitive memory. 
for tractability, we adopt a block-diagonal
approximation (suggested in \cite{wilson2025pop, HansonTuPUP3DGS,HansonSpeedy}) of the information matrix across Gaussian primitives,
\begin{equation}\label{eqn:block_diagonal_approx}
H_{v}:=H_v({\hat{\Theta}})
\approx
\mathrm{blkdiag}(H_v({\hat{\theta}_1}), H_v({\hat{\theta}_2}), \dots, H_v({\hat{\theta}_N})),
\end{equation}
where each block $H_v({\hat{\theta}_i}), $ corresponds to the parameters associated with
an individual Gaussian $i \in \{1, \dots, N\}$.
This approximation preserves intra-splat information while making
storage and incremental updates tractable.

\subsection{View Selection as Binary Optimization}

Let
$\mathcal{V}_c = \mathcal{V} \setminus \mathcal{V}_0  = \{v_1,\dots,v_M\}$
denote a pool of candidate viewpoints available for acquisition.
The problem is to select a subset of informative views that maximally
reduces the uncertainty of the estimated 3DGS parameters. 

Each candidate viewpoint contributes additive
information to the parameter estimate as follows:
\begin{equation}
H_S :=
\sum_{v \in \mathcal{S}} H_{v},
\end{equation}
where $H_v$ denotes the information contribution associated with
candidate viewpoint $v$, and
$\mathcal{S} \subseteq \mathcal{V}_c$
represents the selected set of views.

To quantify uncertainty reduction, we consider a scalar uncertainty metric
$\psi(\cdot)$
defined over the accumulated information matrix.
The view selection problem can therefore be written as
\begin{equation}\label{eqn:cov_minimization}
\min_{S \subseteq \mathcal{V}_c }
\quad
\psi\!\left(\left(
H_0 + \sum_{v \in S} H_{v}
\right)^{-1}\right)
\quad
\text{s.t.}
\quad
|\mathcal{S}| = K,
\end{equation}
where 
\begin{equation}\label{eq:prior_covariance}
    H_0 \footnote{\textrm{To address potential ill-conditioning, we precondition the matrix as $H_0 \leftarrow H_0 + \delta I$, where $\delta > 0$ is a small regularization parameter. }} := \Sigma_0^{-1} = \sum_{v_{0} \in \mathcal{V}_0} H_{v_{0}} = \sigma^{-2}\sum_{v_{0} \in \mathcal{V}_0}
J_{v_{0}}^\top J_{v_{0}},
\end{equation}
denotes the prior block diagonal information matrix, obtained from training 3DGS with initial training set and $K$ is the view cardinality, denoting the number of views to be selected in a batch.
Equivalently, introducing binary decision variables
$x_v \in \{0,1\}$ indicating whether a candidate viewpoint is selected,
the view selection problem can be expressed as
\begin{equation}\label{eqn:binary_optimization}
\min_{{\mathrm{x}_v} \in \{0,1\}}
\,
\psi\!\left(
\left(
H_0
+
\sum_{v\in\mathcal{V}_c}
{\mathrm{x}_v}  H_v
\right)^{-1}\right)
\quad
\text{s.t.}
\quad
\sum_{v\in\mathcal{V}_c}^{} {\mathrm{x}_v}  = K .
\end{equation}

The resulting problem is a combinatorial optimization over binary variables and is generally intractable to solve exactly. This motivates the use of efficient approximation algorithms.

\section{Methodology}\label{sec:method}
The binary optimization problem formulated in \eqref{eqn:binary_optimization} is typically solved using greedy algorithms.
However, classical greedy strategies \cite{nemhauser1978maximizing_submodular_Set_functions} require exhaustive evaluation over the entire candidate set, resulting in substantial latency. 

To address this, we propose LiTe-GS, an oracle-efficient next best view selection framework
for 3D Gaussian Splatting. Specifically,
we leverage stochastic submodular optimization techniques \cite{mirzasoleiman2015stochastic_greedy} to design an efficient active-view selection strategy capable of balancing estimation accuracy with oracle efficiency.


An overview of the proposed pipeline and its comparison with POP-GS \cite{wilson2025pop} is shown in Fig.~\ref{fig:pipeline}. LiTe-GS retains
the same Fisher-information-based view-selection objective as POP-GS and
modifies only the candidate evaluation within the view-selection module.
Specifically, the Fisher-information evaluation of a candidate view serves
as the oracle, and LiTe-GS reduces the number of such oracle evaluations
through randomized candidate sampling. The subsequent 3DGS optimization is
unchanged. Thus, our efficiency analysis concerns the information-driven
view-selection module, rather than end-to-end 3DGS training time.

\begin{algorithm}[t]
\caption{LiTe-GS: Oracle-Efficient NBV Selection}
\label{alg:litegs}

\begin{algorithmic}[1]

\State \textbf{Input:}
$H_0$, candidate views $\mathcal V_c$ with $|\mathcal V_c|=M$,
view cardinality $K$, accuracy parameter $\varepsilon$

\State \textbf{Output:} Selected set $\mathcal S$

\State $\mathcal S \leftarrow \emptyset$, \quad $H \leftarrow H_0$

\State Initialize marginal gains  $\gamma_v \leftarrow \infty$ for all $v\in\mathcal V_c$

\For{$q=1$ to $K$}
    \State Sample subset
    $\mathcal P\subset(\mathcal V_c\setminus\mathcal S)$
    of size
    $p=\frac{M}{K}\log\!\left(\frac{1}{\varepsilon}\right)$
    \State Initialize max-priority queue $\mathcal Q$
    over $v\in\mathcal P$ using \hspace*{0.5cm} gains $\gamma_v$
    \Repeat
        \State Pop view $v$ with the largest cached gain from $\mathcal Q$
        \State \textbf{Oracle evaluation:}
        recompute the true \hspace*{1.1cm}marginal gain
        \[
        \Delta_v=
        \rho(\{v\}\mid\mathcal S)
        \]
        \State Update cached gain        \(
        \gamma_v\leftarrow\Delta_v
        \)
        \State Re-insert $v$ into $\mathcal Q$ with updated priority $\gamma_v$
    \Until{$\gamma_v=\max_{u\in\mathcal P}\gamma_u$}
    \State $\mathcal S\leftarrow\mathcal S\cup\{v\}$, \quad $H\leftarrow H+H_v$
\EndFor

\State \Return $\mathcal S$

\end{algorithmic}
\end{algorithm}

\subsection{Uncertainty Metric}
Uncertainty in the
estimated Gaussian parameters is characterized through the posterior
covariance matrix. A natural measure of uncertainty for multivariate
Gaussian distributions is the differential entropy\cite{Cover2006}, which is a monotonically increasing function of the log-determinant of its covariance matrix.  This objective function in \eqref{eqn:cov_minimization} can be written as:
\begin{equation}
\psi(\mathcal{S})
=
-\log \det
\left(
H_0 + \sum_{v \in \mathcal{S}} H_{v}
\right).
\label{eqn:utility_function}
\end{equation}
 To measure the reduction in uncertainty relative to the prior, we instead define the set function:
\begin{equation}
\rho(\mathcal S)
=
\psi(\emptyset)
-
\psi(\mathcal S).
\label{eqn:normalized_utility}
\end{equation}
By construction, $\rho(\emptyset)=0$. We define the marginal gain of adding a candidate view $v$ to a set $\mathcal{S}_q \subset \mathcal{S}$ as:
\begin{equation}
\rho(\{v\}|\mathcal{S}_q)
:=
\rho(\mathcal{S}_q \cup \{v\}) - \rho(\mathcal{S}_q).
\label{eqn:marginal_gain}
\end{equation}

We define an \emph{oracle evaluation} as the computation of
$\rho({v}\mid\mathcal{S}_q)$ for a candidate view $v$. This computation
constructs and accumulates the corresponding Fisher-information
contribution and is the basic computational unit analyzed in our
information-driven view-selection framework.

\subsection{Performance Analysis}

We analyze the performance of Algorithm~\ref{alg:litegs}
with respect to oracle complexity and theoretical optimality
guarantees.

\subsubsection{Oracle Efficiency Analysis}

The classical greedy algorithm for submodular maximization requires
$O(MK)$ oracle evaluations to select $K$ views from a candidate pool
of size $M$, since all remaining candidates must be evaluated at each
iteration. 

In contrast, Algorithm~\ref{alg:litegs} employs
randomized sampling together with lazy marginal evaluation.
At each iteration, only
\(
p=\frac{M}{K}\log\!\left(\frac{1}{\varepsilon}\right)
\)
candidate views are evaluated. Over $K$ iterations, the total number
of oracle evaluations becomes
\begin{equation}\label{eq:num_oracle_eval}
 O\!\left(M \log \frac{1}{\varepsilon}\right),   
\end{equation}
which is independent of the view cardinality $K$.

\subsubsection{Approximation Guarantees}

We now characterize the approximation performance of
Algorithm~\ref{alg:litegs} relative to the optimal
solution $\mathcal{S}^*$ of the optimization problem
in~\eqref{eqn:cov_minimization}.

\begin{theorem}
\label{thm:stochastic_greedy_performance}
Let $\rho(\cdot)$ be a normalized, non-negative, monotone submodular function.
For a sampling size
$p=\frac{M}{K}\log\!\left(\frac{1}{\varepsilon}\right)$,
Algorithm~\ref{alg:litegs} achieves
\begin{equation}
\mathbb{E}\!\left[\rho(\mathcal{S})\right]
\geq
\left(
1 - e^{-(1-\varepsilon)}
\right)
\rho(\mathcal{S}^*),
\label{eqn:stochastic_greedy_guarantee}
\end{equation}
using only
$O\!\left(M \log \frac{1}{\varepsilon}\right)$
oracle evaluations.
\end{theorem}

Theorem~\ref{thm:stochastic_greedy_performance} establishes that the
proposed method achieves a near-optimal solution in expectation while
maintaining oracle complexity independent of cardinality $K$, and depending only on accuracy parameter $\varepsilon$. 
The full proof and detailed analysis of Theorem~\ref{thm:stochastic_greedy_performance} are provided in the supplementary material.


Most importantly, $\varepsilon$ provides a tunable parameter that explicitly controls the trade-off between oracle efficiency and approximation quality. Increasing $\varepsilon$ reduces the number of oracle evaluations in~\eqref{eq:num_oracle_eval} at the expense of a weaker approximation guarantee in Theorem~\ref{thm:stochastic_greedy_performance}. In contrast, POP-GS performs exhaustive oracle evaluations and therefore does not offer such a trade-off.

\section{Experiments} \label{sec:exp}

\begin{figure*}[t]
\centering

\begin{subfigure}{0.20\textwidth}
    \centering
    \includegraphics[width=\linewidth]{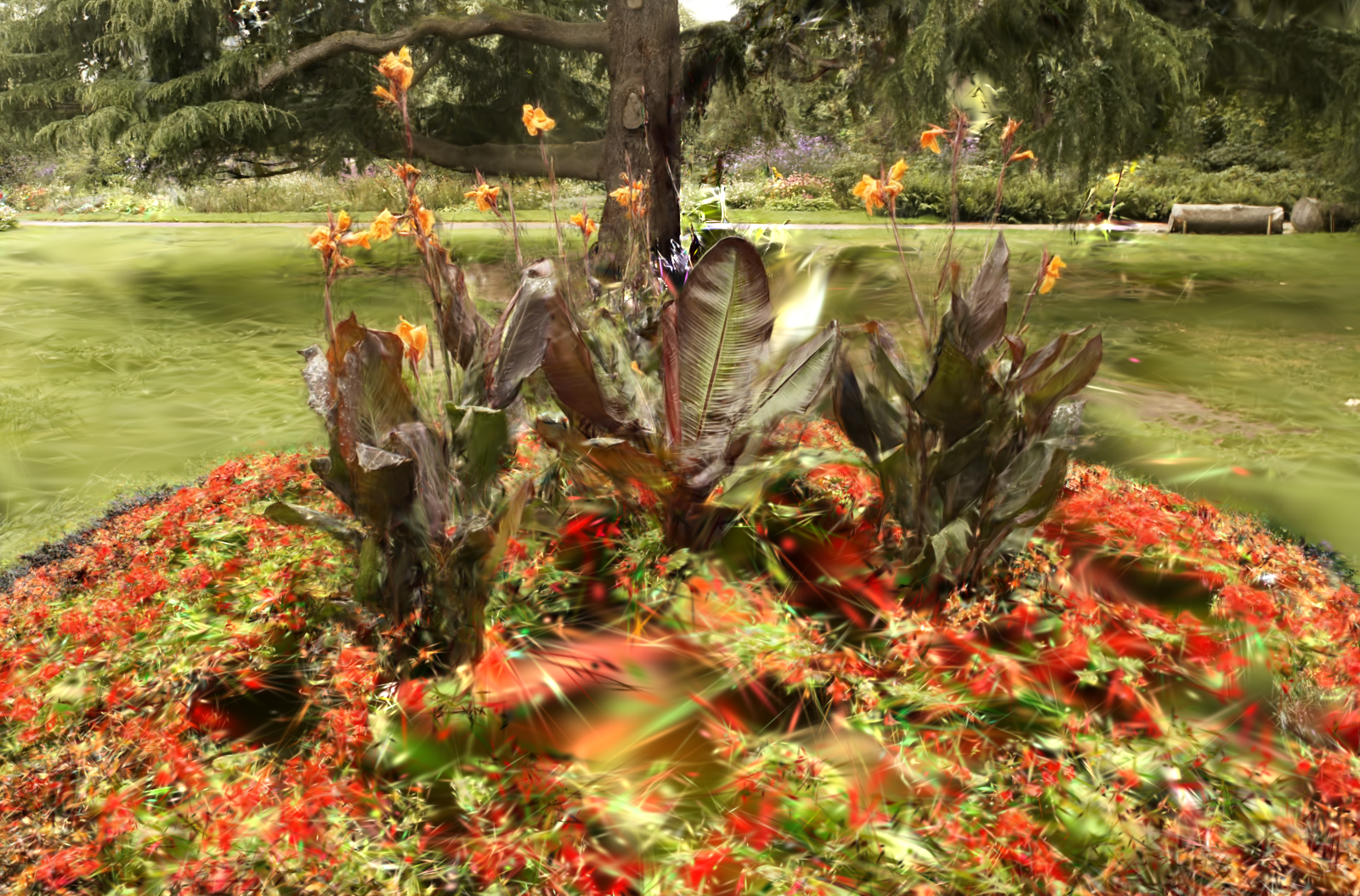}
\end{subfigure}
\begin{subfigure}{0.20\textwidth}
    \centering
    \includegraphics[width=\linewidth]{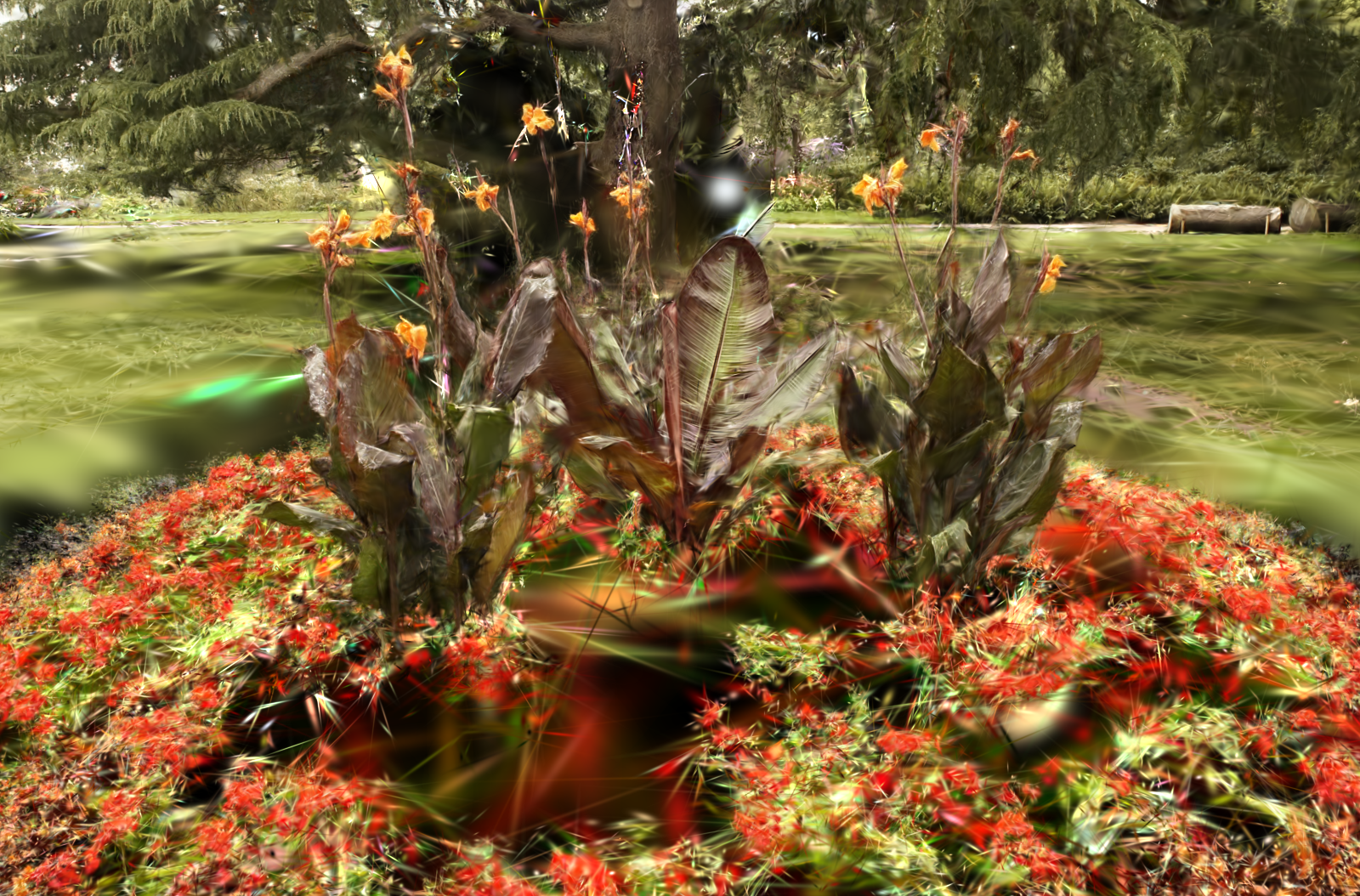}
\end{subfigure}
\begin{subfigure}{0.20\textwidth}
    \centering
    \includegraphics[width=\linewidth]{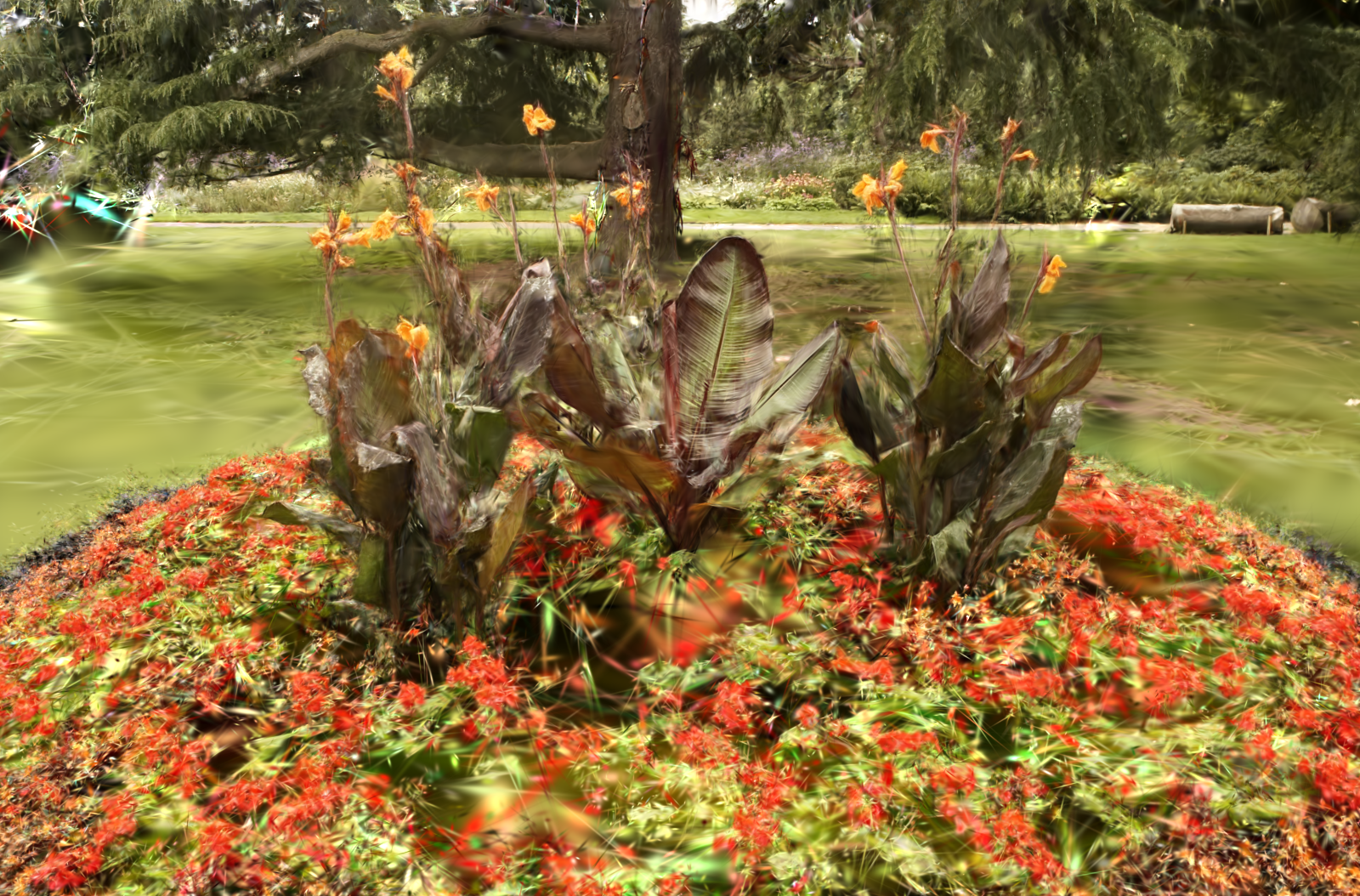}
\end{subfigure}
\begin{subfigure}{0.20\textwidth}
    \centering
    \includegraphics[width=\linewidth]{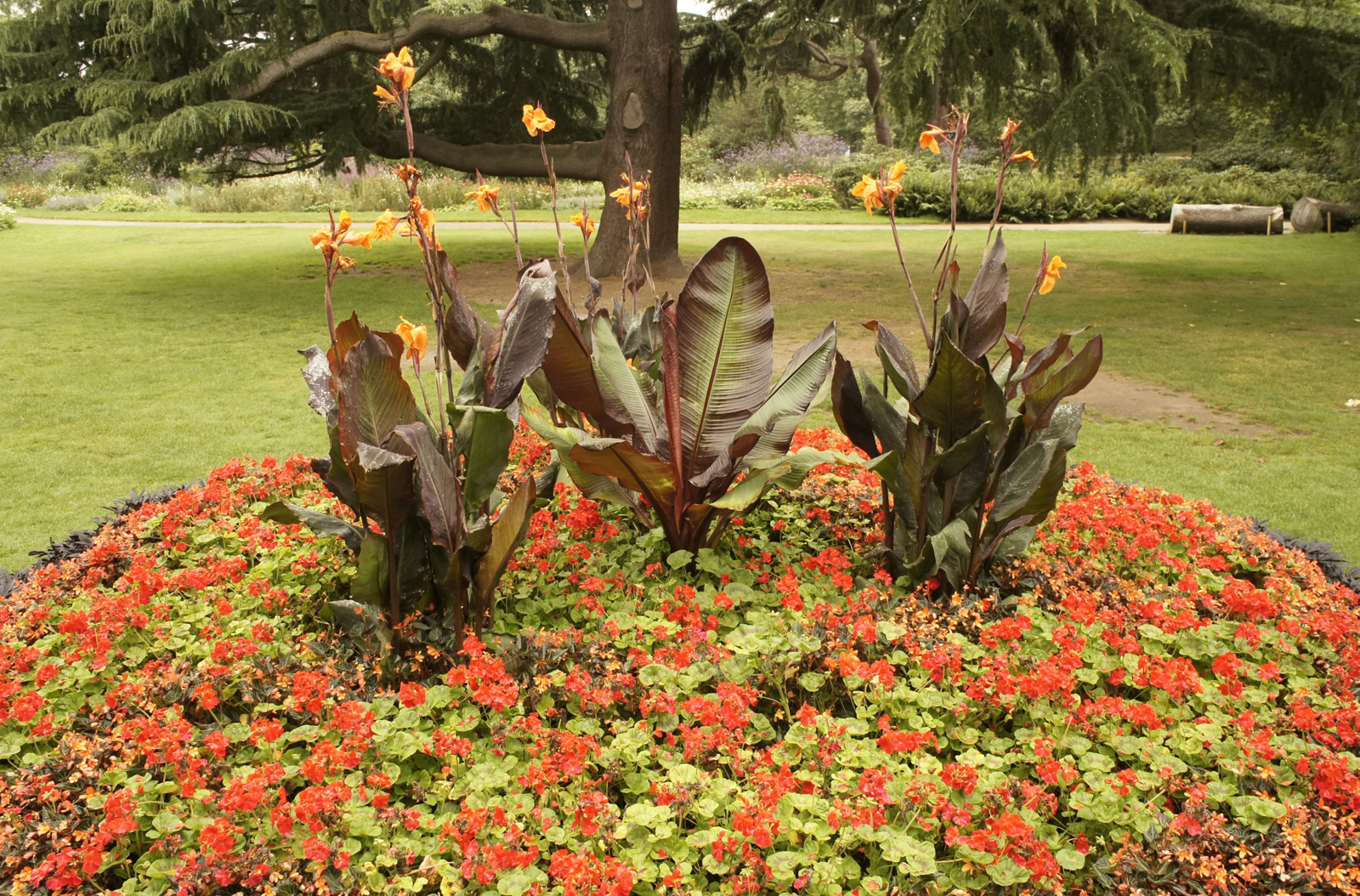}
\end{subfigure}

\vspace{0.1cm}

\begin{subfigure}{0.20\textwidth}
    \centering
    \includegraphics[width=\linewidth]{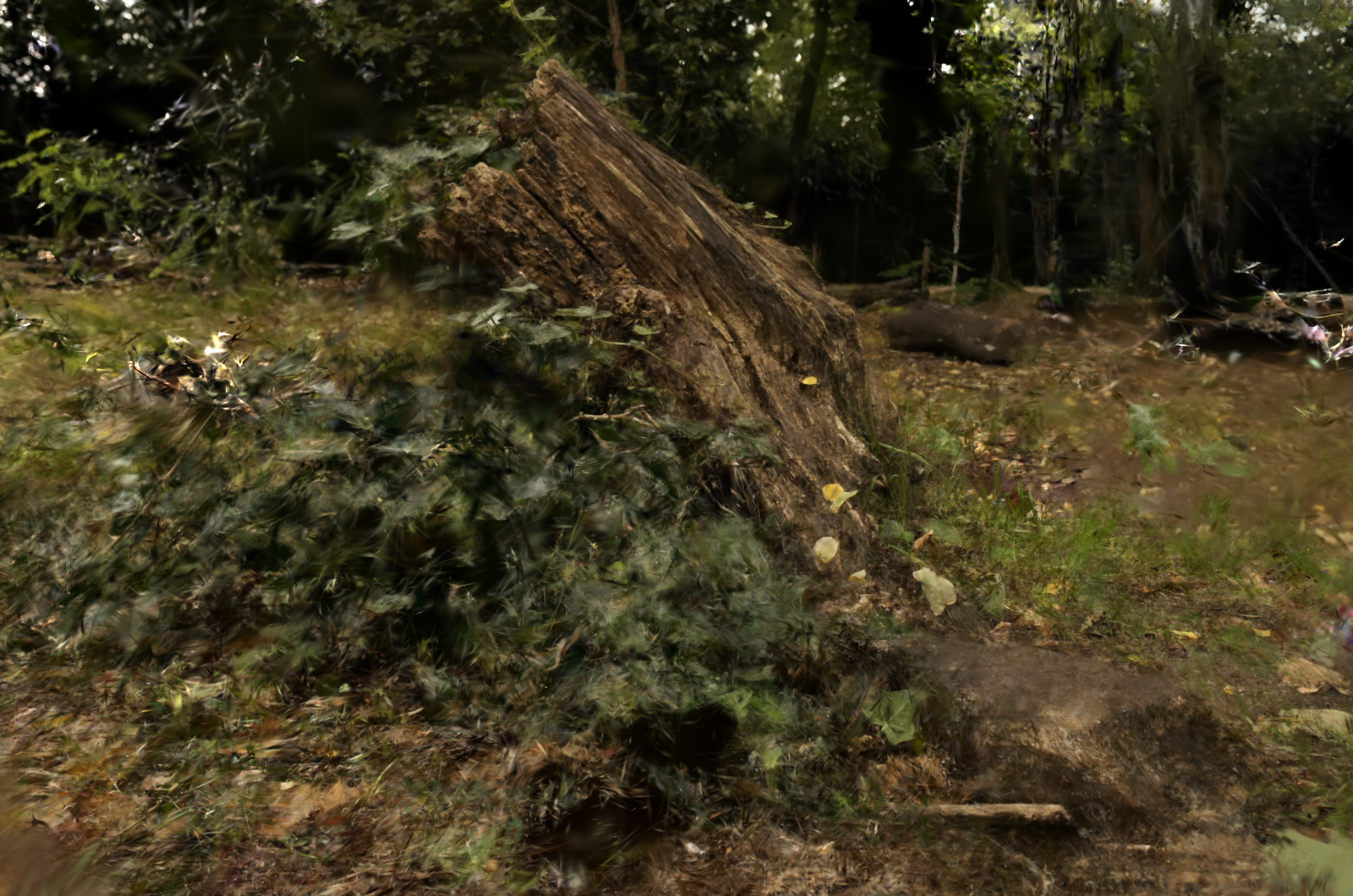}
\end{subfigure}
\begin{subfigure}{0.20\textwidth}
    \centering
    \includegraphics[width=\linewidth]{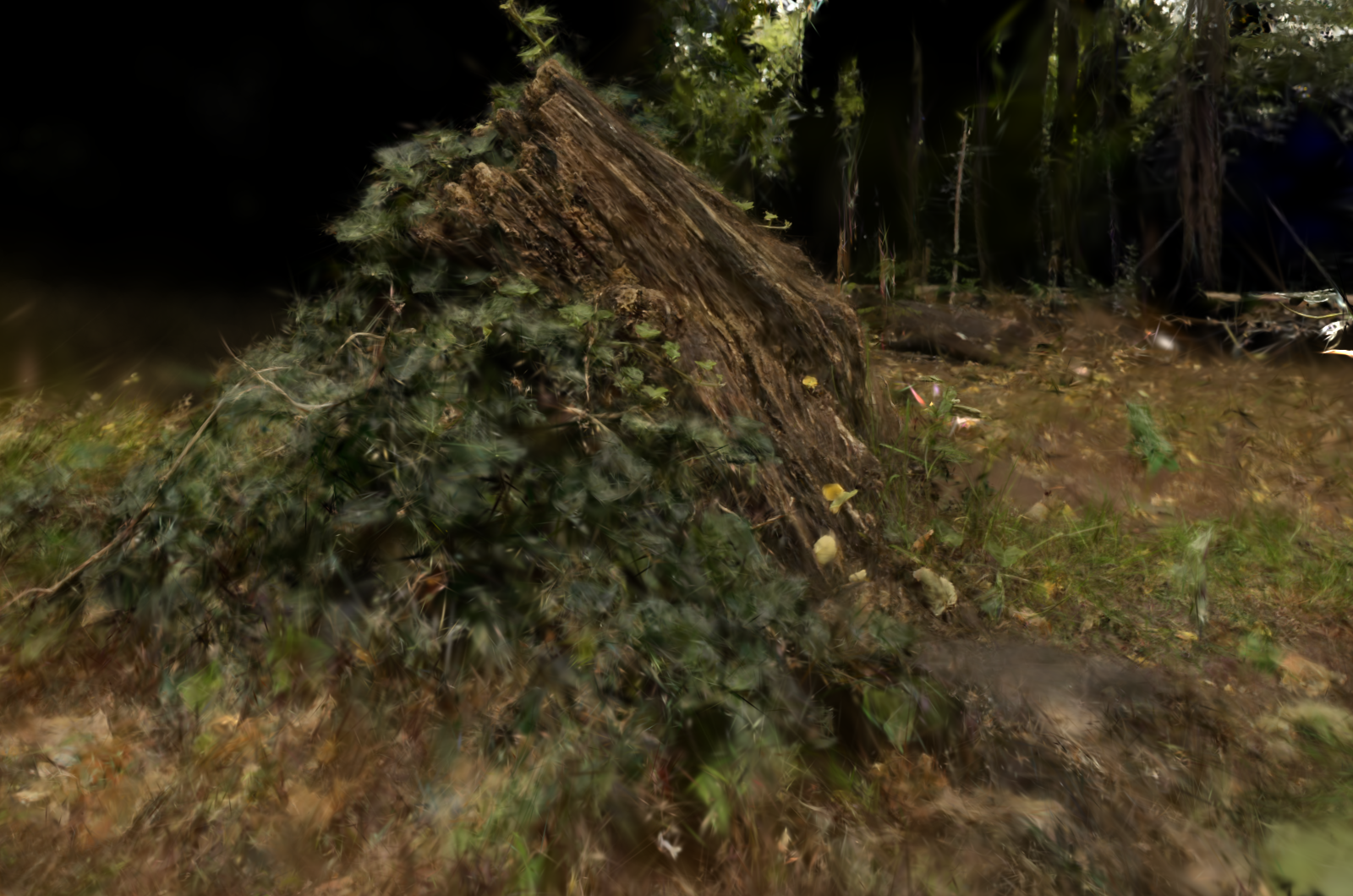}
\end{subfigure}
\begin{subfigure}{0.20\textwidth}
    \centering
    \includegraphics[width=\linewidth]{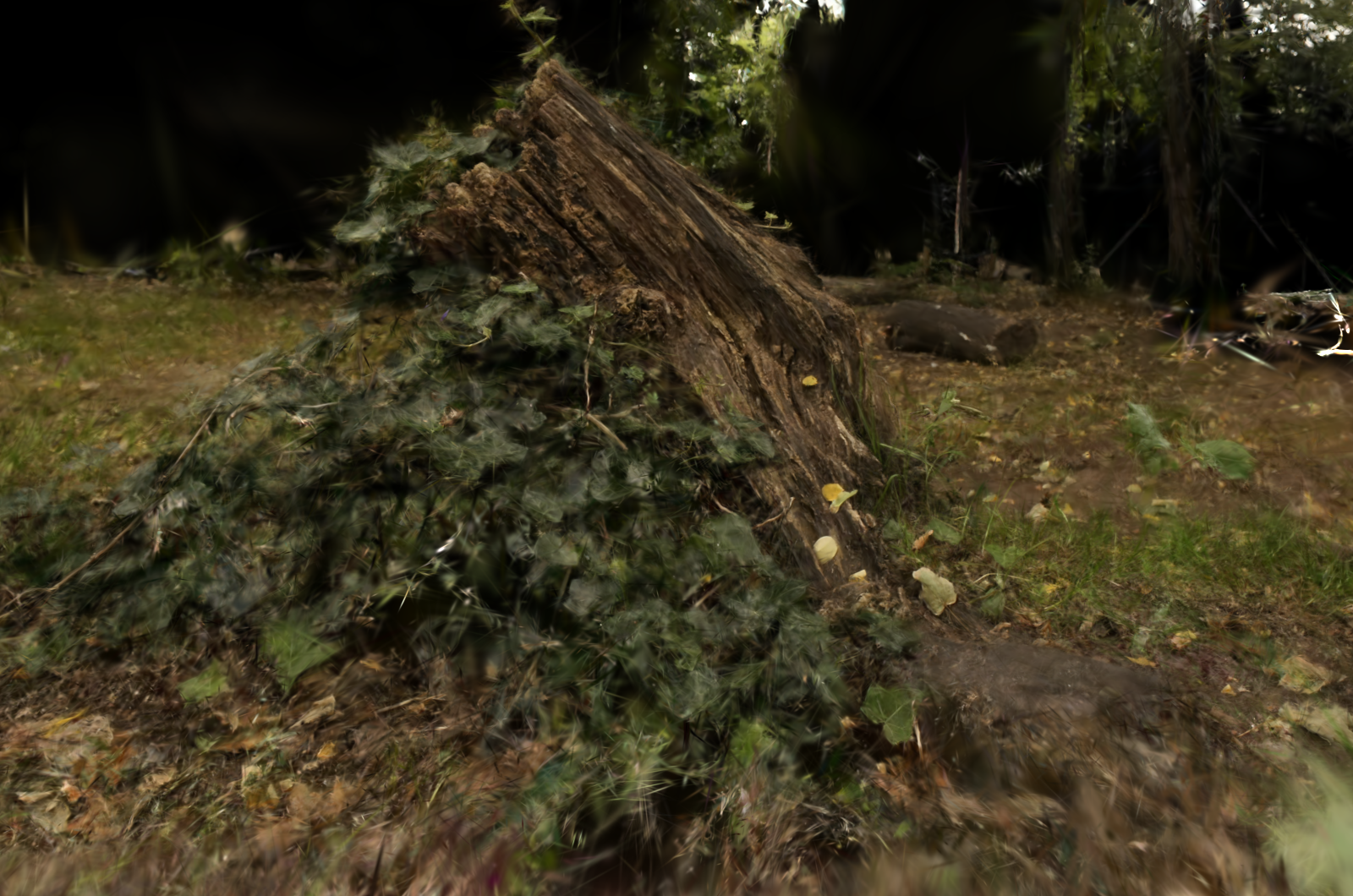}
\end{subfigure}
\begin{subfigure}{0.20\textwidth}
    \centering
    \includegraphics[width=\linewidth]{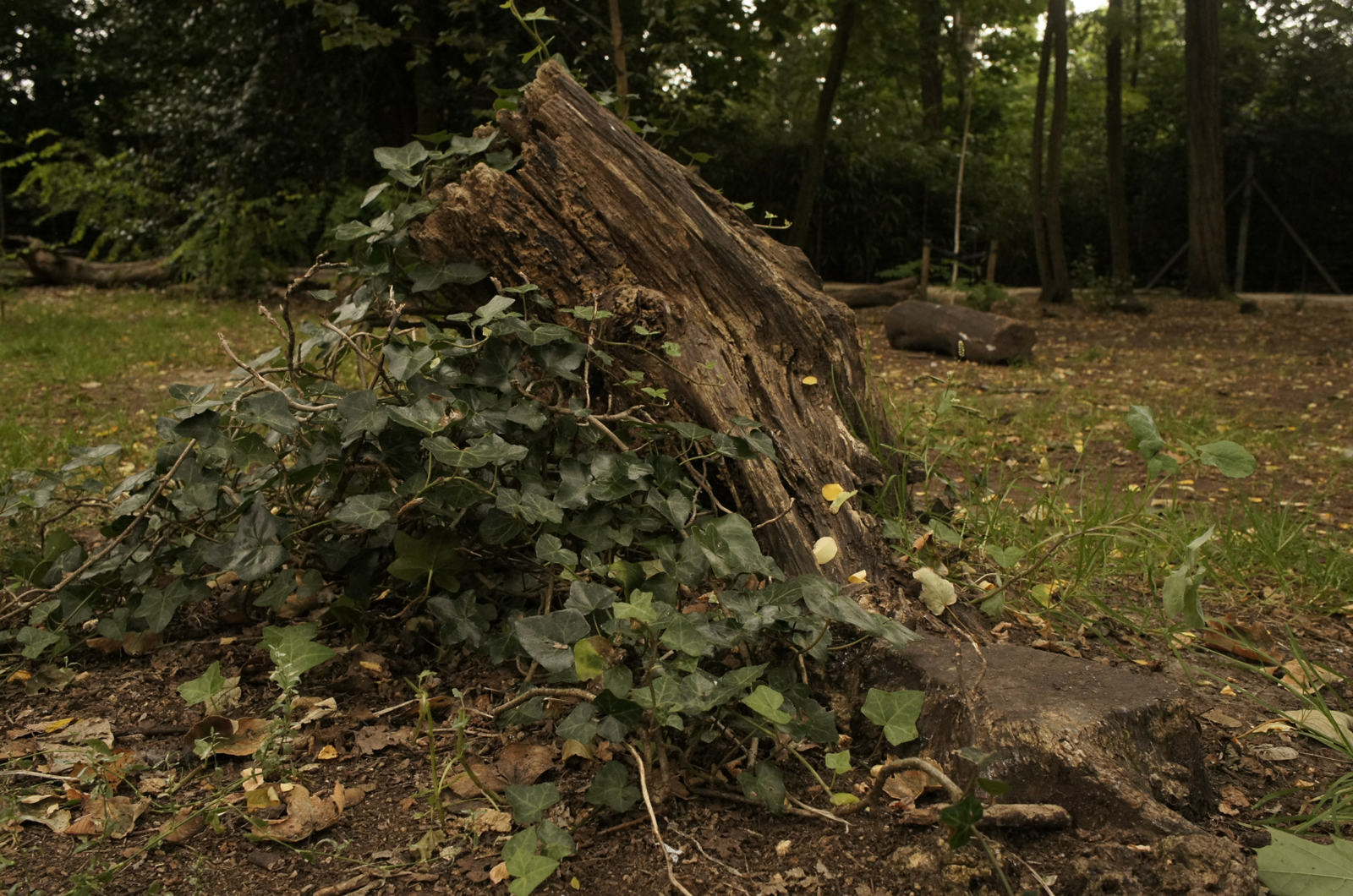}
\end{subfigure}

\vspace{0.1cm}

\begin{subfigure}{0.20\textwidth}
    \centering
    \includegraphics[width=\linewidth]{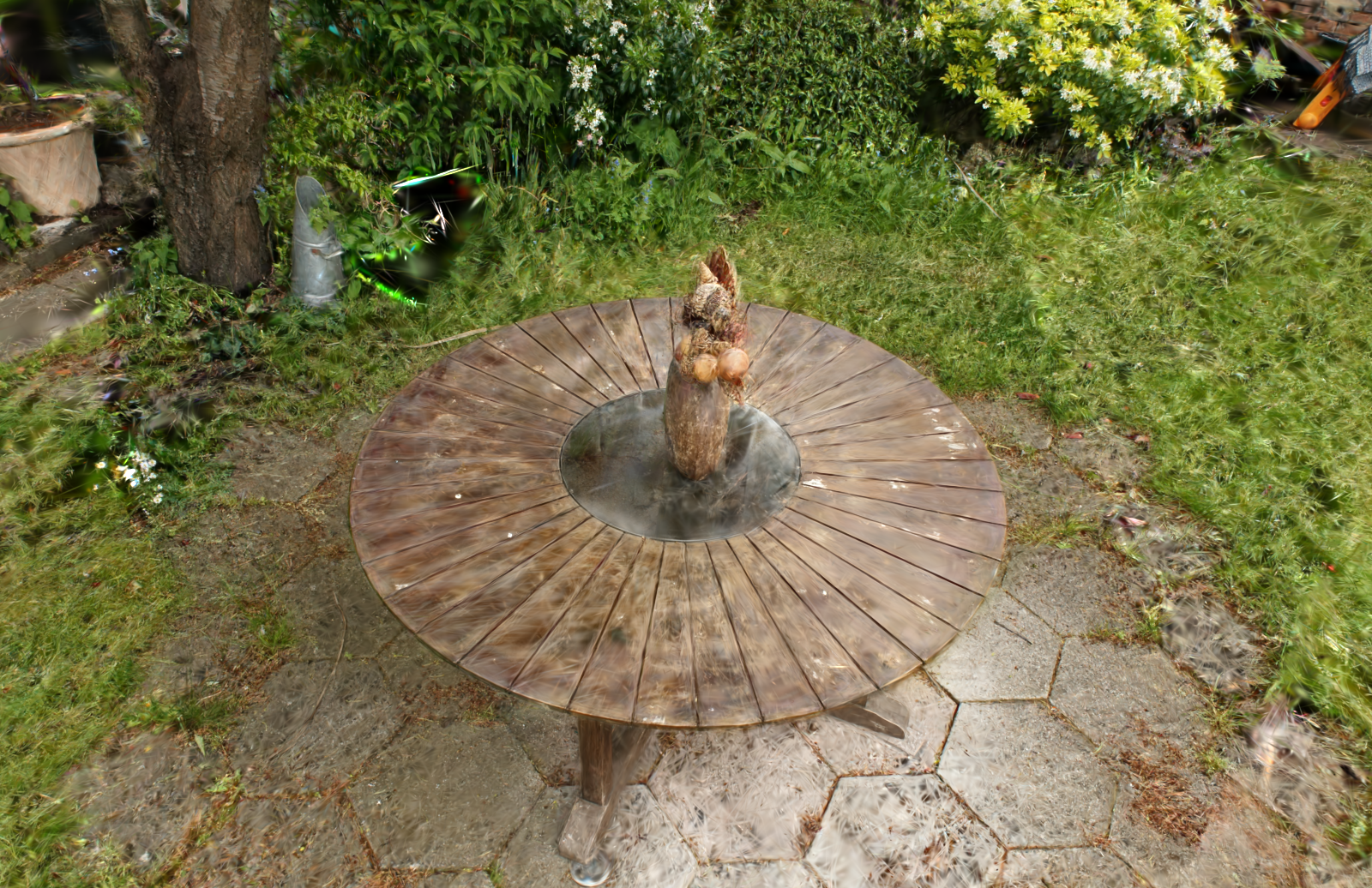}
    \caption{Random}
\end{subfigure}
\begin{subfigure}{0.20\textwidth}
    \centering
    \includegraphics[width=\linewidth]{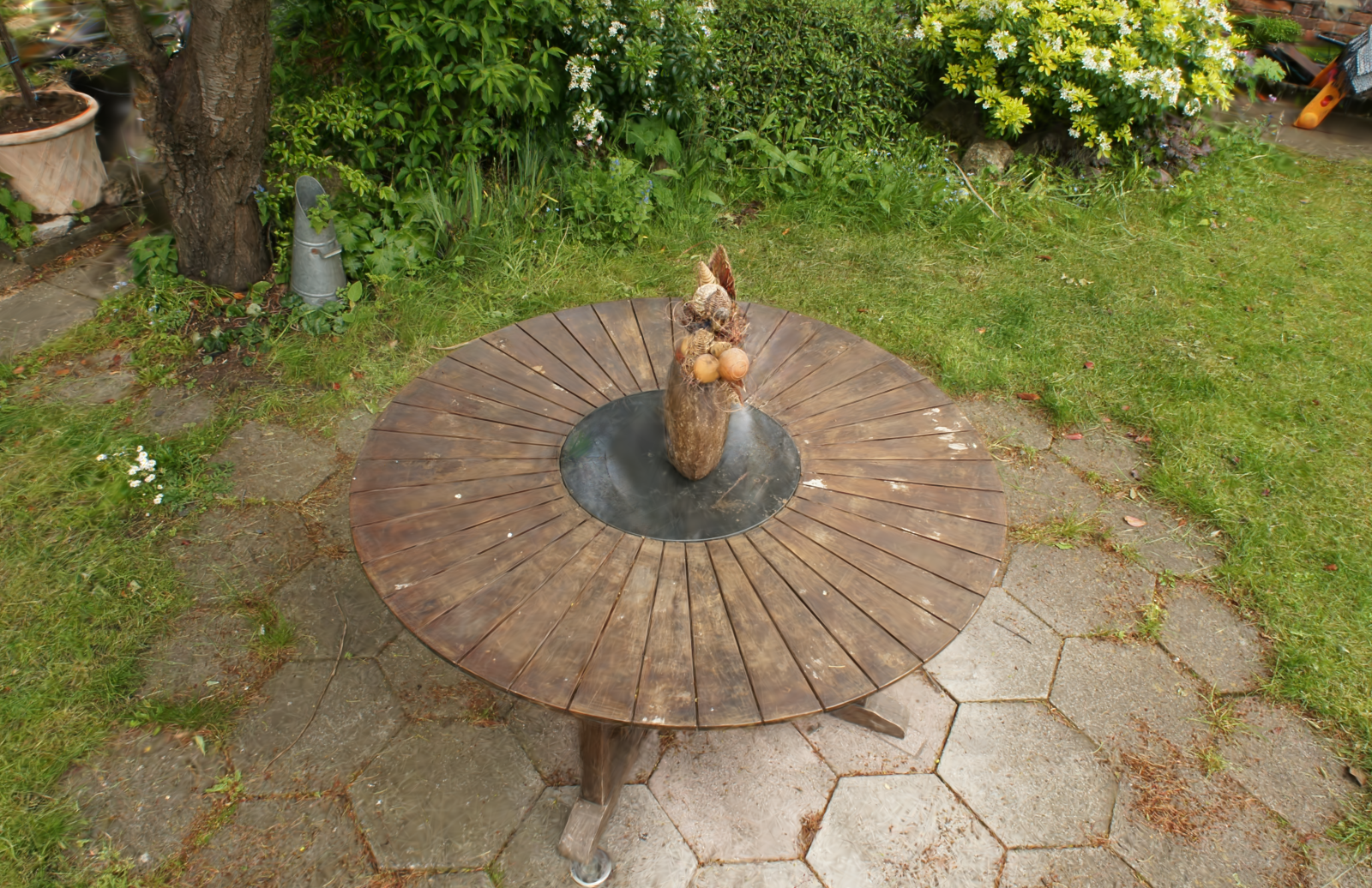}
    \caption{POP-GS}
\end{subfigure}
\begin{subfigure}{0.20\textwidth}
    \centering
    \includegraphics[width=\linewidth]{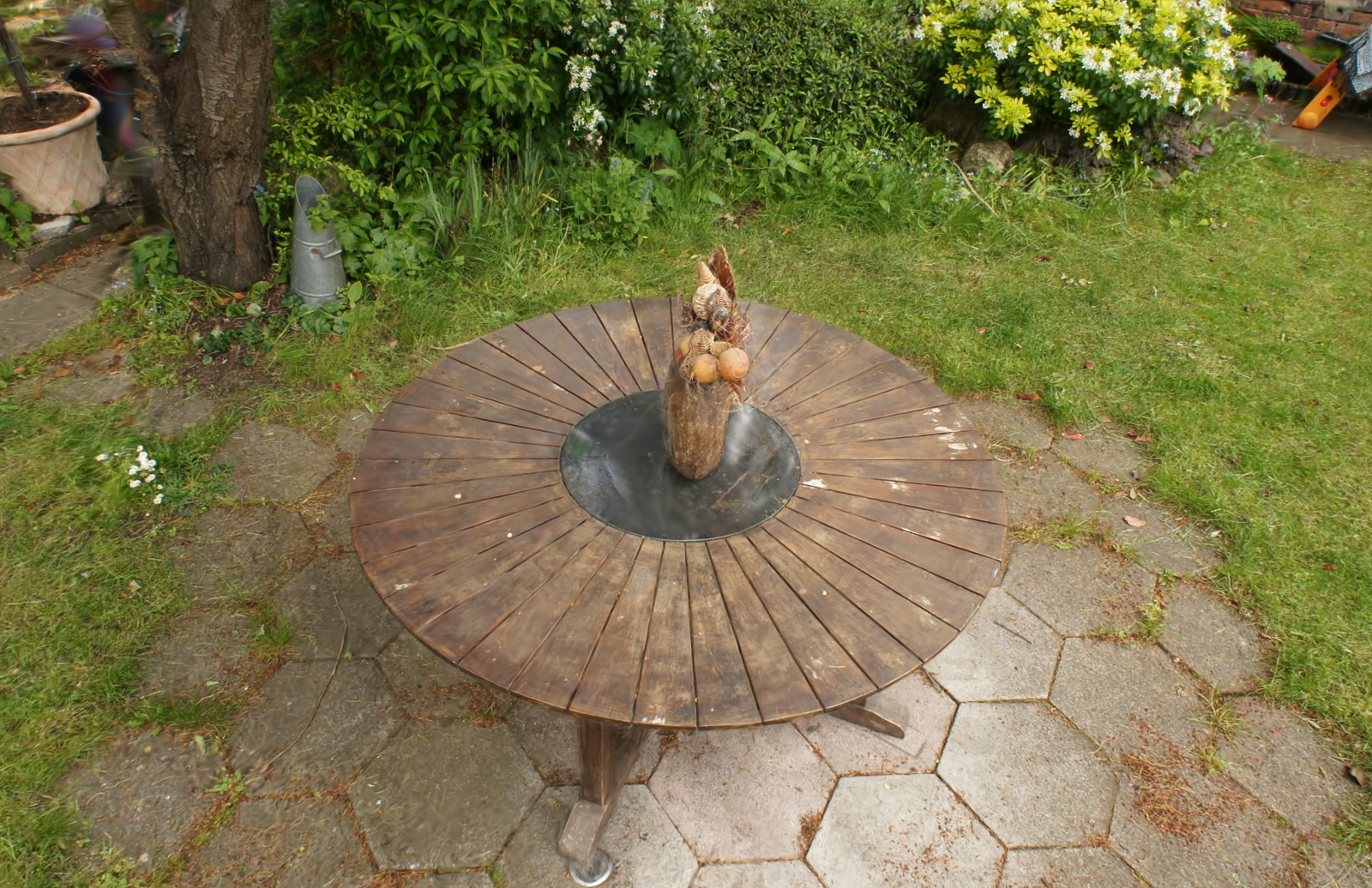}
    \caption{Ours}
\end{subfigure}
\begin{subfigure}{0.20\textwidth}
    \centering
    \includegraphics[width=\linewidth]{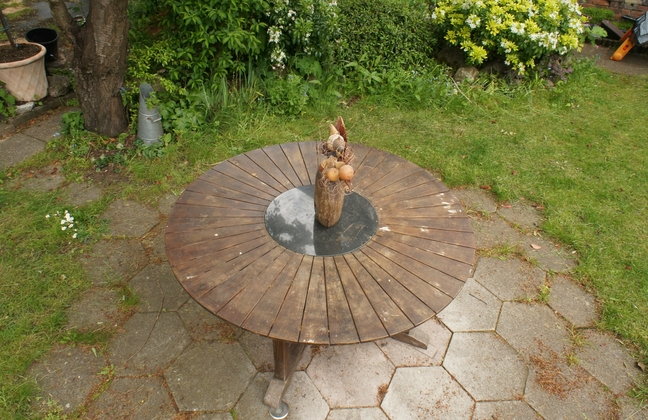}
    \caption{Ground Truth}
\end{subfigure}
\caption{Setup 1: Qualitative comparison of rendered images on the Mip-NeRF 360 dataset. Results obtained using Random sampling, POP-GS~\cite{wilson2025pop}, and LiTe-GS are shown alongside the corresponding ground-truth images.}
\label{fig:mipnerf_exp1}
\vspace{-0.25cm}
\end{figure*}
  
In this section, we evaluate the proposed LiTe-GS algorithm, describing the dataset, baseline comparisons, experimental setup, and evaluation metrics, followed by a detailed analysis of both qualitative and quantitative results. For all experiments, we set $\varepsilon = 0.1$ in Algorithm~\ref{alg:litegs}, unless otherwise stated. Our experimental framework is built upon the FisherRF repository \cite{Jiang2023FisherRF}, which we extend to incorporate our proposed algorithm, including the block-diagonal approximation of the Fisher Information \eqref{eqn:block_diagonal_approx}.

\begin{table}[t]
\centering
\footnotesize
\renewcommand{\arraystretch}{1.21}
\setlength{\tabcolsep}{2.51pt}
\begin{tabular}{l|ccc|ccc}
\toprule
\textbf{Methods} 
& \multicolumn{3}{c|}{\textbf{Setup 1}} 
& \multicolumn{3}{c}{\textbf{Setup 2}} \\ 
\cmidrule(lr){2-4}  \cmidrule(lr){5-7} 
& {PSNR ↑} & {SSIM ↑} & {LPIPS ↓}
& {PSNR ↑} & {SSIM ↑} & {LPIPS ↓} \\
\midrule
Random     & 19.52 & 0.830 & 0.150 & 24.42  & 0.893 & 0.096  \\
POP-GS       & 22.10 & 0.857 & 0.118  & 25.25  & 0.901  & 0.090   \\
Ours     & 21.97 & 0.856 & 0.119  & 25.31  & 0.902 & 0.089   \\
\bottomrule
\end{tabular}
\caption{
Quantitative Comparison of rendered images across Blender scenes. Metrics are averaged over the three evaluated scenes. 
}
\label{tab:blender_exp1}
\vspace{-0.5cm}
\end{table}

\subsection{Experimental Setup and Dataset}
We evaluate the proposed approach on two widely used benchmarks for novel view synthesis and 3D scene reconstruction. The first is the Blender dataset~\cite{mildenhall2022nerf}, which consists of eight synthetically rendered scenes, each containing 100 training images and 200 test images. The second is the Mip-NeRF 360 dataset~\cite{barron2022mip}, which comprises high-resolution real-world scenes with significantly greater geometric and visual complexity. Together, these datasets enable evaluation across both controlled synthetic environments and realistic scenes.

We compare our method with the D-Optimality criterion implemented POP-GS~\cite{wilson2025pop}, along with random selection as baselines. 
We do not include FisherRF~\cite{Jiang2023FisherRF} as a separate baseline because it uses a diagonal Fisher approximation. Under the diagonal approximation, the POP-GS formulation reduces to the FisherRF formulation, making FisherRF a special case of the same Fisher-information framework.
The comparison metrics are {Peak Signal-to-Noise Ratio (PSNR)}~\cite{gonzalez2008digital}, {Structural Similarity Index (SSIM)}~\cite{wang2004image}, and {Learned Perceptual Image Patch Similarity (LPIPS)}~\cite{zhang2018perceptual}. To assess the proposed method under different operating conditions, we consider the following three experimental setups:

\subsubsection{Setup 1} 
The training process is initialized with a seed set of two images. Subsequently, three images are selected and added at each iteration until the training set contains a total of 11 images. After each acquisition step, the model is trained for \(100v\) iterations, where \(v\) denotes the current number of training images. The total training step consists of 11,000 training iterations.

\subsubsection{Setup 2} 
The training process is initialized with five images. At each acquisition step, four additional images are selected until the training set reaches 21 images. Between acquisition steps, the model is trained for \(300v\) iterations, where \(v\) is the current training-set size. The total training step consists of 21,000 training iterations. This experiment is conducted exclusively on the Blender dataset.

\subsubsection{Setup 3}
The training process is initialized with three images. Subsequently, four images are added per acquisition step until the training set reaches 15 images. After each acquisition step, the model is trained for \(100v\) iterations. The total training step consists of 11,000 training iterations, and is conducted exclusively on the Mip-NeRF 360 dataset.

\subsection{Results }

We now present both qualitative and quantitative evaluations of the proposed method \footnote{Please refer to the supplementary material for detailed and additional results.}.

\subsubsection{Rendered Image Quality}

Figures~\ref{fig:blender_exp1} and \ref{fig:mipnerf_exp1} show qualitative comparisons of the rendered images against ground truth for Blender and Mip-NeRF360 dataset respectively. Visually, our method produces reconstructions that are highly comparable to those obtained with POP-GS.
This observation is further supported by the quantitative results reported in Tables~\ref{tab:blender_exp1} and \ref{tab:mipnerf_exp1} for Blender and Mip-NeRF 360 dataset respectively. Each table reports average performance over the three scenes considered for each dataset. Across most scenes, LiTe-GS achieves performance comparable to POP-GS in terms of PSNR, SSIM, and LPIPS, while consistently outperforming random selection, which serves as a lower-bound baseline.

\begin{table}[t]
\centering
\footnotesize
\renewcommand{\arraystretch}{1.21}
\setlength{\tabcolsep}{2.51pt}
\begin{tabular}{l|ccc|ccc}
\toprule
\textbf{Methods} 
& \multicolumn{3}{c|}{\textbf{Setup 1}} 
& \multicolumn{3}{c}{\textbf{Setup 3}} \\ 
\cmidrule(lr){2-4}  \cmidrule(lr){5-7} 
& {PSNR ↑} & {SSIM ↑} & {LPIPS ↓}
& {PSNR ↑} & {SSIM ↑} & {LPIPS ↓} \\
\midrule
Random     & 16.71 & 0.381 & 0.461 & 17.68  & 0.414 & 0.437  \\
POP-GS       & 17.72 & 0.449 & 0.420  & 18.82  & 0.502  & 0.397   \\
Ours     & 17.54 & 0.446 & 0.423  & 19.01  & 0.506 & 0.398   \\
\bottomrule
\end{tabular}
\caption{
Quantitative comparison of rendered images across Mip-NeRF 360 scenes. Metrics are averaged over the three evaluated scenes. 
}
\label{tab:mipnerf_exp1}
\vspace{-0.5cm}
\end{table}

In some cases, our method even outperforms POP-GS. This improvement is particularly interesting given that our algorithm evaluates only a randomly sampled subset of candidate views at each iteration. While deterministic greedy selection always makes the locally optimal choice, such decisions may be suboptimal from a global perspective. In contrast, the randomized selection strategy introduces controlled stochasticity, which can occasionally escape locally optimal yet globally inferior selections. Although LiTe-GS is not expected to systematically outperform POP-GS in
reconstruction quality, occasional improvements can arise from stochastic
candidate sampling. These cases illustrate the potential benefit of
exploring different candidate subsets, rather than constituting a claimed
quality advantage over POP-GS.



Next, we analyze the oracle efficiency of LiTe-GS.

\subsubsection{Oracle Efficiency Results}
We measure the efficiency of information-driven view selection by the
number of Fisher-information (oracle) evaluations. This oracle-based
measure is independent of hardware and implementation details and directly
quantifies the number of information queries required by the selection
algorithm.


We compare LiTe-GS against POP-GS in terms of the percentage reduction in
oracle evaluations. For each dataset, the reduction is computed relative to
POP-GS, and the final reported value is obtained by averaging across
datasets. As shown in Table~\ref{tab:oracle_efficiency}, LiTe-GS achieves substantial
reductions in the number of Fisher oracle evaluations across all experimental
settings. This reduction directly reflects the effectiveness of the proposed
subset-based evaluation strategy, which avoids exhaustive scoring of all
candidate views at each iteration.
In contrast to POP-GS, which evaluates all candidate views at every iteration,
the proposed method restricts oracle evaluations to a sampled subset whose
size depends on the accuracy parameter $\varepsilon$, thereby improving
oracle efficiency.
For conservative oracle accounting, we count all
sampled candidates as oracle evaluations, without crediting the
additional savings from lazy evaluation.

\begin{table}[t]
\centering
\footnotesize
\renewcommand{\arraystretch}{1.21}
\setlength{\tabcolsep}{5.51pt}
\begin{tabular}{c c c c}
\toprule
Setup & POP-GS & LiTe-GS & Reductions(\%)\\
\midrule

1(Blender) & 282
& 215
& 23.76 \\

1(Mip-NeRF 360) & 403
& 300
& 25.56 \\


2 & 350
& 200
& 42.86 \\


3 & 522
& 302
& 42.15 \\

\bottomrule
\end{tabular}
\caption{Oracle efficiency comparison across different view acquisition setups.
We report the number of Fisher information (oracle) evaluations for POP-GS and LiTe-GS,
along with the corresponding percentage reduction achieved by LiTe-GS relative to POP-GS.}
\label{tab:oracle_efficiency}
\vspace{-0.21cm}
\end{table}

\begin{table*}[t]
\centering
\footnotesize
\renewcommand{\arraystretch}{1.21}
\setlength{\tabcolsep}{5.51pt}
\begin{tabular}{cc|cccc|cccc|c}
\toprule
\multicolumn{2}{c|}{}
& \multicolumn{4}{c|}{\textbf{POP-GS}}
& \multicolumn{4}{c|}{\textbf{LiTe-GS}}
& \\
\textbf{$K$} & \textbf{$\varepsilon$}
& \textbf{PSNR $\uparrow$} & \textbf{SSIM $\uparrow$}
& \textbf{LPIPS $\downarrow$} & \textbf{\# Oracle}
& \textbf{PSNR $\uparrow$} & \textbf{SSIM $\uparrow$}
& \textbf{LPIPS $\downarrow$} & \textbf{\# Oracle}
& \textbf{Oracle Reduction}
\\
\midrule

3 & 0.1
& 25.63 & 0.832 & 0.162 & 275
& 25.43 & 0.834 & 0.160 & 210
& 23.64\% \\

4 & 0.1
& 24.63 & 0.823 & 0.164 & 366
& 25.49 & 0.830 & 0.158 & 209
& 42.90\% \\

\multirow{3}{*}{6}
& 0.1
& \multirow{3}{*}{25.45}
& \multirow{3}{*}{0.830}
& \multirow{3}{*}{0.154}
& \multirow{3}{*}{549}
& 25.07 & 0.820 & 0.160 & 208
& 62.11\% \\

& 0.3
&&&&
& 24.62 & 0.824 & 0.160 & 108
& 80.42\% \\

& 0.6
&&&&
& 23.75 & 0.811 & 0.169 & 44
& 91.99\% \\

\bottomrule
\end{tabular}
\caption{Effect of selection cardinality $K$ and approximation parameter
$\varepsilon$ on oracle efficiency and reconstruction quality on
Blender (Ship scene). The first three rows fix $\varepsilon=0.1$ and vary $K$;
the last three rows fix $K=6$ and vary $\varepsilon$.}
\label{tab:ship_vs_K}
\end{table*}

We additionally evaluate the dependence of LiTe-GS on the selection cardinality $K$ and $\varepsilon$ in \textbf{Setup 4}. Training is initialized with three images, followed by $K\in \{3,4,6\}$ images added at each acquisition step until reaching 15 images; each stage uses $300v$ training iterations, for a total of 15,000 iterations. As shown in Table~\ref{tab:ship_vs_K}, increasing $K$ from 3 to 6 increases POP-GS oracle evaluations from 275 to 549, whereas LiTe-GS remains nearly constant (210, 209, and 208), increasing the oracle reduction from 23.64 \% to 62.11 \%. Thus, the experiment directly demonstrates the predicted $K$-independence of LiTe-GS oracle complexity. For fixed $K=6$, increasing $\varepsilon$ further reduces LiTe-GS oracle evaluations from 208 to 108 and 44, yielding 80.42\% and 91.99\% reductions, respectively, while incurring the expected approximation-quality trade-off.

\begin{table}[t]
\centering
\footnotesize
\renewcommand{\arraystretch}{1.21}
\setlength{\tabcolsep}{2.51pt}
\begin{tabular}{l|ccc|ccc}
\toprule
 & \multicolumn{3}{c|}{\textbf{LiTe-GS}} & \multicolumn{3}{c}{\textbf{SSG}} \\ 
\cmidrule(lr){1-1} \cmidrule(lr){2-4} \cmidrule(lr){5-7}
$\quad \varepsilon $  & {PSNR ↑} & {SSIM ↑} & {LPIPS ↓} & {PSNR ↑} & {SSIM ↑} & {LPIPS ↓} \\
\midrule
$0.60 ~~ $     & \textbf{24.35} & \textbf{0.819} & \textbf{0.166} & 22.11  & 0.805 & 0.181  \\
$0.40 ~~ $     & \textbf{24.18} & {0.816} & {0.168} & 24.11  & \textbf{0.821} & \textbf{0.166}  \\
$0.25 ~~ $     & {25.04} & {0.823} & {0.162} & \textbf{25.40}  & \textbf{0.829} & \textbf{0.158}  \\
$0.20 ~~ $       & \textbf{25.47} & {0.827} & {0.159}  & 25.17  & \textbf{0.832}  & \textbf{0.158}   \\
$0.10 ~~ $     & \textbf{25.49} & \textbf{0.830} & {0.158}  & 25.38  & {0.829} & 0.158   \\
\bottomrule
\end{tabular}
\caption{
Effect of dynamic randomized resampling on the Blender (Ship scene).
LiTe-GS dynamically resamples candidate subsets at each acquisition step,
whereas Static-Subset Greedy (SSG) uses a fixed subset sampled once per
stage. Both use the same subset size determined by $\varepsilon$. The training protocol
follows Setup~4 with $K=4$.
}
\label{tab:blender_ablation}
\end{table}


\subsection{Ablation Study}

We study the effect of dynamic randomized resampling by comparing LiTe-GS
with a Static-Subset Greedy (SSG) baseline. For each $\varepsilon$, both
methods use the same subset size determined by the sampling rule; LiTe-GS
resamples the candidate subset at each acquisition step, whereas SSG
samples once per stage and keeps the subset fixed. Thus, the comparison
directly isolates dynamic versus static candidate restriction.

Table~\ref{tab:blender_ablation} reports results on Blender
(\textit{Ship}). A fixed subset can occasionally be favorable, allowing
SSG to outperform LiTe-GS for a particular setting. This is expected:
SSG can benefit from a favorable subset realization, but remains
permanently restricted to it. LiTe-GS instead continues to explore
different candidate subsets at each acquisition step, reducing dependence
on any single subset realization.
Table~\ref{tab:mip_nerf_ablation} reports the same comparison on
Mip-NeRF~360 (\textit{Garden}), where LiTe-GS consistently outperforms SSG
across the evaluated $\varepsilon$ values. Together, the two studies illustrate that dynamic
resampling is not guaranteed to improve reconstruction quality for every
individual realization, but provides a more robust candidate-exploration
mechanism than permanent restriction to a fixed subset.

The ablation also illustrates the accuracy-efficiency trade-off
characterized by Theorem~\ref{thm:stochastic_greedy_performance}:
increasing $\varepsilon$ reduces candidate evaluations while permitting a
controlled approximation-quality trade-off. Unlike exhaustive POP-GS
selection, LiTe-GS therefore provides an explicit mechanism for tuning
oracle expenditure to the desired accuracy level.

\begin{table}[t]
\centering
\footnotesize
\renewcommand{\arraystretch}{1.21}
\setlength{\tabcolsep}{2.51pt}
\begin{tabular}{l|ccc|ccc}
\toprule
 & \multicolumn{3}{c|}{\textbf{LiTe-GS}} & \multicolumn{3}{c}{\textbf{SSG}} \\ 
\cmidrule(lr){1-1} \cmidrule(lr){2-4} \cmidrule(lr){5-7}
$\quad \varepsilon$  & {PSNR ↑} & {SSIM ↑} & {LPIPS ↓} & {PSNR ↑} & {SSIM ↑} & {LPIPS ↓} \\
\midrule
$0.60 ~~ $     & \textbf{20.42} & \textbf{0.563} & \textbf{0.407} & 19.82  & 0.533 & 0.437  \\
$0.50 ~~ $       & \textbf{20.37} & \textbf{0.558} & \textbf{0.406}  & 20.25  & 0.548  & 0.433   \\
$0.40 ~~ $     & \textbf{20.33} & \textbf{0.558} & \textbf{0.406}  & 20.17  & 0.541 & 0.436   \\
$0.25 ~~ $     & \textbf{20.50} & {0.555} & \textbf{0.408}  & 19.92  & 0.555 & 0.410   \\
\bottomrule
\end{tabular}
\caption{
Effect of dynamic randomized resampling on the Mip-NeRF 360 (Garden scene).
LiTe-GS dynamically resamples candidate subsets at each acquisition step,
whereas Static-Subset Greedy (SSG) uses a fixed subset sampled once per
stage. Both use the same subset size determined by $\varepsilon$. Training is initialized with two images, followed by three images added at each acquisition step until reaching 11 images; each stage uses $300v$ training iterations, for a total of 78,00 iterations.
}
\label{tab:mip_nerf_ablation}
\end{table}

\section{Conclusion}\label{sec:conclusion}

We presented LiTe-GS, an oracle-efficient active view selection framework
for 3D Gaussian Splatting. We formulate next-best-view selection as cardinality-constrained maximization of Fisher-information gain, enabling efficient approximate view selection.
 We proposed a
randomized strategy, which reduces the
number of oracle evaluations by restricting computation to sampled
subsets at each iteration. This results in efficient selection with
provable approximation guarantees.
Experimental results demonstrate that LiTe-GS achieves reconstruction
quality comparable to state-of-the-art,
while significantly reducing the number of
Fisher information oracle evaluations.
Overall, LiTe-GS provides an efficient framework for
active view selection in 3D Gaussian Splatting.

\newpage
{
    \small
    \bibliographystyle{ieeenat_fullname}
    \bibliography{main}
}
\end{document}